\pdfoutput=1
\documentclass[11pt]{article}

\usepackage[letterpaper,margin=1in]{geometry}
\usepackage{times}
\usepackage[round,authoryear]{natbib}
\usepackage{authblk}

\usepackage[utf8]{inputenc}
\usepackage[T1]{fontenc}
\usepackage{amsmath,amssymb,amsthm,mathtools}
\usepackage{graphicx}
\usepackage{hyperref}
\usepackage{xcolor}
\usepackage{booktabs}
\usepackage{siunitx}
\usepackage{dcolumn}
\newcolumntype{d}[1]{D{.}{.}{#1}}

\usepackage{subcaption}
\usepackage{float}
\usepackage{microtype}

\usepackage{tcolorbox}
\tcbuselibrary{skins, breakable}
\usepackage{tikz}
\usetikzlibrary{arrows.meta, decorations.pathreplacing}

\newcounter{algline}
\newenvironment{pseudocode}[1]{%
  \setcounter{algline}{0}%
  \begin{tcolorbox}[
    title={\small\textbf{#1}},
    fonttitle=\bfseries,
    colback=gray!6,
    colframe=black!55,
    boxrule=0.5pt,
    arc=3pt,
    left=6pt, right=6pt, top=4pt, bottom=4pt,
    breakable
  ]%
  \small\sffamily
  \newcommand{\pcline}[1]{%
    \stepcounter{algline}%
    \noindent\makebox[1.8em][r]{\textcolor{black!40}{\scriptsize\thealgline}}\hspace{0.5em}##1\par\vspace{1pt}%
  }%
  \newcommand{\pckw}[1]{\textbf{##1}}%
  \newcommand{\pccomment}[1]{\hfill\textcolor{black!50}{\(\triangleright\) \textit{##1}}}%
  \newcommand{\ind}{\hspace{1.5em}}%
  \newcommand{\indd}{\hspace{3.0em}}%
  \newcommand{\inddd}{\hspace{4.5em}}%
  \newcommand{\pcsep}{\vspace{2pt}\hrule\vspace{3pt}}%
}{%
  \end{tcolorbox}%
}

\newcommand{\R}{\mathbb{R}}
\newcommand{\E}{\mathbb{E}}
\newcommand{\N}{\mathcal{N}}

\newcommand{\norm}[1]{\left\|#1\right\|}
\newcommand{\given}{\,|\,}

\newtheorem{theorem}{Theorem}
\newtheorem{lemma}{Lemma}
\newtheorem{proposition}{Proposition}

\hypersetup{colorlinks=true, linkcolor=blue!60!black, citecolor=blue!60!black,
            urlcolor=blue!60!black}

\title{Restrained Diffusion Bridges: Targeted Denoiser Training and Refinement}
\author[1]{Xuanxi Zhang}
\author[2]{Kevin S. Cheng}
\author[2]{Weizhou Wang}
\author[2]{Aaron R. Dinner\thanks{\texttt{dinner@uchicago.edu}}}
\author[1]{Jonathan Weare\thanks{\texttt{weare@nyu.edu}}}
\affil[1]{Courant Institute School of Mathematics, Computing, and Data Science, New York University, New York, New York 10012, United States}
\affil[2]{Department of Chemistry and James Franck Institute, University of Chicago, Chicago, Illinois 60637, United States}
\date{}

\begin{document}
\maketitle

\begin{abstract}
Diffusion models are often inaccurate in critical, data-sparse regions of the target distribution. Sampling from restrained distributions, biased distributions used in classical enhanced sampling to concentrate data near chosen configurations, can control where training data are collected but biases the learned distribution.
We establish that this bias cancels exactly when forward noising from restrained data is conditioned on the restrained configuration at an appropriately chosen intermediate time.
This allows a pretrained model fine-tuned on restrained data to achieve high accuracy in any chosen region of the target distribution, regardless of its probability, using standard denoising or flow matching objectives without importance weights. When no  target data are available, Gibbs sampling enables inference from a model trained exclusively on restrained data.
\end{abstract}

\section{Introduction}
\label{sec:intro}

A fundamental challenge in diffusion-based generative modeling is poor accuracy in regions of state space that have low probability under the (possibly unnormalized) target measure $\pi$ and are  poorly represented in training data. These regions, such as transition states in molecular simulations, rare posterior configurations, or low-frequency image modes, are often of critical importance in downstream analysis. A potential remedy is  to reweight the training distribution toward rare regions or collect more data there. However, simple reweighting or oversampling causes the model to learn the wrong target measure. Correcting with importance weights, when they are available, does not resolve the problem because those weights are small in low probability regions by design.

Classical umbrella sampling uses harmonic biases to restrain simulations near
selected configurations  \citep{torrie1977umbrella,umbrella_sampling}.
The resulting densities $\pi(x)\exp(-\|x-z\|^2/2s^2)$ suppress distant
configurations and keep a chain near $z$, facilitating sampling even when $z$
is highly atypical under $\pi$.
As for reweighting and oversampling, a diffusion model trained directly on restrained data learns the restrained density rather than $\pi$.

Here, we show that conditioning the forward noising process on restrained data exactly cancels the biasing effect of those data. A model trained on restrained data with standard denoising or flow matching objectives, and no other modifications, therefore recovers the unconditional score of $\pi$ with no importance weights or correction terms (Theorem~\ref{thm:training}).
This cancellation exploits the  observation that the restrained density at scale $s$
coincides exactly with the posterior distribution $\pi_{0\mid t^*}(x_0\given x_{t^*})$
inside a diffusion model when the noise level $t^*$ is matched to $s$
(Lemma~\ref{lem:posterior}).

\begin{figure}
    \centering
\includegraphics[width=\textwidth]
  {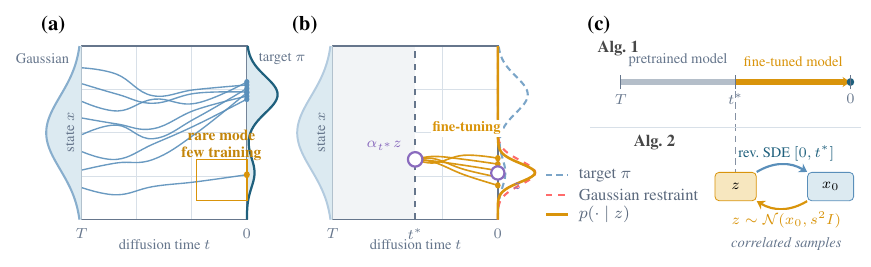}
    \caption{Conceptual overview of Restrained Diffusion Bridges (RDBs).
\textbf{(a)} Standard training provides many trajectories to the dominant mode
of $\pi$ but few to the rare mode.
\textbf{(b)} Multiplying $\pi$ (blue dashed) by the Gaussian restraint (red
dashed) gives the restrained target $p(\cdot\mid z)$ (orange).  Restrained
endpoints and bridges pinned at $x_{t^*}=\alpha_{t^*}z$ focus fine-tuning on the
rare region while recovering the score of $\pi$.
\textbf{(c)} Algorithm 1 uses the pretrained model from $T$ to $t^*$ and the
fine-tuned model from $t^*$ to $0$.  Algorithm 2 alternates the reverse bridge
with $z\sim\mathcal{N}(x_0,s^2I)$ using only the fine-tuned model}
\label{fig:conceptual_bridge_three_panel_compressed}
\end{figure}

These results motivate a targeted training algorithm that requires no additional samples from $\pi$ and allows precise control over where training
data are concentrated without introducing any bias (Figure~\ref{fig:conceptual_bridge_three_panel_compressed}). It can also be used to fine-tune a pretrained diffusion model whose accuracy degrades near a rare configuration: the model is refined only for times in $[0,t^*]$ and only in a neighborhood of that configuration, using restrained samples drawn near it. When $\nabla\log\pi$ is available, we sample the restrained distributions using Langevin sampling.
We also propose an alternative dataset resampling strategy when only samples from $\pi$ are available (Appendix~\ref{app:dataset_resampling}).
The fine-tuned
model produces independent samples via standard inference at no extra cost.

When no reasonable pretrained model is available, our  algorithm can be used to train a model valid on $[0,t^*]$ only. The model on $[0,t^*]$ can then be used to generate samples from $\pi$ using
a learned Gibbs sampler (Section~\ref{sec:gibbs}).
We validate our approach on the Müller-Brown potential, where analytic ground
truth is available, and on two molecular benchmarks, alanine dipeptide in explicit
solvent and chignolin in implicit solvent, which test the approach in realistic
simulation settings (Section~\ref{sec:experiments}).

\section{Related work}
\label{sec:related}

Our goal is unconditional sampling from $\pi$. Diffusion posterior
sampling \citep{Chung2023} and classifier or classifier-free guidance
\citep{Dhariwal2021,Ho2022} condition or steer inference without refining the
marginal score in selected regions. Methods that learn samplers from an unnormalized target include path-space variational and stochastic-control objectives \citep{zhang2022path,Vargas2023,berner2024optimal,richter2024improved,DomingoEnrich2024,sendera2024improved,Havens2025} and iterative energy matching \citep{AkhoundSadegh2024}. These methods use $\pi$ or
$\nabla\log\pi$ in the loss. Our work uses the target only to generate restrained data, which are then used for training with an ordinary denoising loss. It therefore also applies to
implicit marginals, and its dataset-resampling variant requires samples but no
energy evaluations (Appendix~\ref{app:dataset_resampling}).

Restricted-Gaussian conditionals underlie proximal Gibbs sampling \citep{Lee2021} and related diffusion-assisted MCMC methods \citep{Kim2023,Chen2024}.
\citet{chen2026markov}  implement a  Gibbs transition through learned denoising and a  Metropolis correction to remove score and discretization bias.
Generative Gibbs methods
\citep{Wang2026GenerativeGibbsPI,Wang2026GenerativeGibbsPA,Wang2026Nuclear}  exploit the freedom to
learn a conditional $p(x\mid y)$ under a convenient marginal over $y$ and
compose it with a different $y$-distribution at inference. In those papers that freedom is used to reduce the dimensionality and complexity of the training distribution. Here it is used to focus training on chosen target regions.

Diffusion bridge methods construct endpoint-conditioned generative dynamics \citep{debortoli2021diffusion,Zhou2024}.
In our approach, posterior-matched restrained samples are
bridged to fixed endpoints and the ordinary denoising target recovers the
unconditional score. Endpoint frequencies can
therefore emphasize rare regions. This targeted use of restrained components is
related to stratification, which can improve inverse-temperature scaling of classical sampling for
multimodal targets \citep{Dinner2020}.

\section{Setup}
\label{sec:setup}

We use  $x_t$ for forward noised  states.
We reserve bare $\pi$ for the target density. Subscripted $\pi$ denotes the
finite-dimensional laws induced by the forward process initialized with
$x_0\sim\pi$. For example, $\pi_t$, $\pi_{0,t}$, $\pi_{0\mid t}$, and
$\pi_{t\mid 0,t^*}$ denote the marginal, joint, posterior, and bridge
densities, respectively.

For definiteness, we consider  Gaussian forward processes with marginals
\begin{equation}
  x_t = \alpha_t\, x_0 + \sigma_t\, \varepsilon, \qquad
  \varepsilon \sim \N(0, I), \quad x_0 \sim \pi,
  \label{eq:forward}
\end{equation}
where $\alpha_t>0$, $\sigma_t>0$, and $\alpha_t/\sigma_t$ is strictly
decreasing, with $\alpha_0=1$, $\sigma_0\approx 0$, and
$\alpha_T/\sigma_T\approx 0$ so that $x_T\approx\sigma_T\varepsilon$ is noise-dominated.  Our results generalize beyond Gaussian and continuous space settings with appropriate redefinition of the restrained densities.
We write
$\alpha_{t|s}=\alpha_t/\alpha_s$ for the transition ratio and
$\sigma_{t|s}^2=\sigma_t^2-\alpha_{t|s}^2\sigma_s^2$ for the conditional
variance. The latter is nonnegative when $\alpha_t/\sigma_t$ is
decreasing.
The forward process \eqref{eq:forward} induces the joint and marginal densities
\[
  \pi_{0,t}(x_0,x_t)
  =\pi(x_0)\N(x_t;\alpha_t x_0,\sigma_t^2I),
  \qquad
  \pi_t(x_t)=\int\pi_{0,t}(x_0,x_t)\,\mathrm dx_0,
\]
whose score $\nabla_{x_t}\log \pi_t(x_t)$ drives the reverse dynamics.

The marginal \eqref{eq:forward} is realized by the It\^{o} stochastic differential equation (SDE)
\begin{equation}
  \mathrm{d}x_t
  = \frac{\dot\alpha_t}{\alpha_t}\,x_t\,\mathrm{d}t + g_t\,\mathrm{d}W_t,
  \qquad g_t^2 \;=\; \dot\sigma_t^2 - \frac{2\dot\alpha_t\,\sigma_t^2}{\alpha_t},
  \label{eq:forward_sde}
\end{equation}
where $W_t$ is standard Brownian motion, overdots denote time derivatives,
and $g_t^2\ge 0$ follows from $\alpha_t/\sigma_t$ decreasing. The corresponding reverse-time SDE is
\begin{equation}
  \mathrm{d}x_t
  = \Bigl[\frac{\dot\alpha_t}{\alpha_t}\,x_t
    - g_t^2\,\nabla_{x_t}\!\log \pi_t(x_t)\Bigr]\mathrm{d}t
    + g_t\,\mathrm{d}\bar{W}_t,
  \label{eq:reverse_sde}
\end{equation}
where $\bar{W}_t$ is a reverse-time Brownian motion \citep{Song2021}.  Alternatively, the
probability flow ordinary differential equation (ODE)
\begin{equation}
  \dot{x}_t
  =  f_t^\pi(x_t) , \qquad   f_t^\pi(x_t)
  \;=\; \frac{\dot\alpha_t}{\alpha_t}\,x_t
        - \frac{g_t^2}{2}\,\nabla_{x_t}\!\log \pi_t(x_t),
  \label{eq:pf_ode}
\end{equation}
shares the same marginals as \eqref{eq:reverse_sde} \citep{Song2021} and
generates samples from $\pi$ deterministically.  We refer to both
\eqref{eq:reverse_sde} and \eqref{eq:pf_ode} as the
\emph{reverse dynamics}. The choice between stochastic and deterministic
integration is independent of training.  The variance preserving schedule
($\alpha_t^2+\sigma_t^2=1$) and the flow-matching
schedule ($\alpha_t=1{-}t$, $\sigma_t=t$) \citep{Lipman2023} are special
cases. With the latter schedule, the ODE \eqref{eq:pf_ode} recovers standard flow-matching inference.
A network $\psi_\theta(x_t,t)$ is trained to approximate a chosen
parameterization of either the \emph{score}
$\nabla_{x_t}\!\log \pi_t(x_t)$ or the \emph{velocity field} $ f_t^\pi(x_t)
$
 and then, at inference time, used to drive  the corresponding
reverse dynamics \eqref{eq:reverse_sde} or \eqref{eq:pf_ode} starting from initial conditions sampled from $\N(0, \sigma_T^2 I)$.
Our goal is to either train a new model or refine a pretrained model  $\psi_\theta$ in low-density regions for $t\le t^*$.

We choose a \emph{probe point} $z\in\R^d$ and a \emph{bridge time} $t^*\in(0,T)$,
which together determine the restraint width $s = \sigma_{t^*}/\alpha_{t^*}$.
The restrained density centered at $z$ is
\begin{equation}
  p(x_0 \given z)
  \;\propto\;
  \pi(x_0)\,\exp\!\left(-\frac{\norm{x_0-z}^2}{2s^2}\right).
  \label{eq:restrained}
\end{equation}
Gaussian forward-noise posteriors are common in diffusion-assisted sampling
\citep{Kim2023,Chen2024}. As the following lemma shows, restrained samples at scale $s$ are exact draws from the
diffusion posterior at $x_{t^*}=\alpha_{t^*}z$, the condition under
which bridge score matching recovers the unconditional score of $\pi$
(Theorem~\ref{thm:training}).
\begin{lemma}[Posterior identification]
\label{lem:posterior}
With $s = \sigma_{t^*}/\alpha_{t^*}$, the restrained density equals the
diffusion posterior exactly:
$p(x_0 \given z) = \pi_{0\mid t^*}(x_0 \given \alpha_{t^*}z)$.
\end{lemma}
\begin{proof}
Apply Bayes' theorem to the Gaussian forward kernel, then
substitute $s=\sigma_{t^*}/\alpha_{t^*}$.
\end{proof}

Forward noising $x_t=\alpha_t x_0+\sigma_t\varepsilon$ from
$x_0\sim p(\cdot\given z)$ produces a restrained marginal rather than
$\pi_t$, so ordinary denoising score matching
\citep{Vincent2011,Ho2020} would learn the wrong score. More generally, at a
fixed time $t$, the population denoising regressor depends on the conditional
law $\pi_{0\mid t}(x_0\given x_t)$, not on the marginal distribution from which $x_t$ is
sampled. We could therefore choose any distribution for $x_t$ and draw
$x_0$ from its matching posterior. In the Gaussian case this posterior is a
restrained distribution, but implementing this construction directly would
require a separate restrained simulation for every selected $x_t$ and $t$.

A bridge amortizes these conditional simulations across time. Fix $t^*>t$ and
an endpoint $y$, draw $x_0\sim \pi_{0\mid t^*}(x_0\given y)$, and then draw
$x_t\sim \pi_{t\mid 0,t^*}(x_t\given x_0,y)$. By the Markov property of the forward process,
$\pi_{0\mid t,t^*}(x_0\given x_t,y)=\pi_{0\mid t}(x_0\given x_t).$
Thus ordinary denoising regression has the unconditional population optimum
for every fixed endpoint $y$. The marginal distribution of $x_{t^*}$ may
therefore be chosen arbitrarily, allowing endpoints in rare regions to be
oversampled. Lemma~\ref{lem:posterior} identifies the required endpoint
posterior for $y=\alpha_{t^*}z$ with the restrained ensemble
$p(\cdot\given z)$.

Given a starting point $x_0$ drawn from $p(\cdot\given z)$ and the bridge
endpoint $y = \alpha_{t^*}z$, the forward process conditioned on both values is
a Gaussian bridge:
\begin{equation}
  x_t \given (x_0,\; x_{t^*}=\alpha_{t^*}z)
  \;\sim\;
  \N\!\Bigl(\mu_t(x_0, z),\;\sigma_{\mathrm{br}}^2(t)\,I\Bigr),
  \quad t \in [0, t^*],
  \label{eq:bridge}
\end{equation}
with mean and bridge variance
\begin{equation}
  \mu_t(x_0, z) = \alpha_t\,x_0
    + \frac{\alpha_{t^*|t}\,\alpha_{t^*}\,\sigma_t^2}{\sigma_{t^*}^2}
    \bigl(z - x_0\bigr),
  \qquad
  \sigma_{\mathrm{br}}^2(t) = \frac{\sigma_t^2\,\sigma_{t^*|t}^2}{\sigma_{t^*}^2}
  \;\le\; \sigma_t^2.
  \label{eq:mu}
\end{equation}
Because $\sigma_{\mathrm{br}}^2(t)<\sigma_t^2$ for every
$t\in(0,t^*)$, bridge samples are more concentrated than ordinary forward
samples around the conditional bridge tube connecting $x_0$ to
$\alpha_{t^*}z$. They therefore provide dense training coverage of the
targeted region throughout the low-noise interval.

\section{Restrained bridge-based training}
\label{sec:algorithms}

Algorithm~1 trains a diffusion model on $[0,t^*]$ from samples of the restrained density $p(\cdot\given z)$.
To generate training data we assume access to the true score $\nabla_{x_0}\!\log\pi(x_0)$ (e.g.,\ the
negative gradient of a potential energy), so the restrained score is
$\nabla_{x_0}\log p(x_0 \given z) = \nabla_{x_0}\log\pi(x_0) - (x_0 - z)/s^2$.
Overdamped Langevin dynamics with this score confines the chain to a neighborhood around $z$, regardless of how atypical $z$ is under $\pi$, at
the same per-step cost as Langevin on $\pi$ itself (one call to
$\nabla_{x_0}\!\log\pi$). An alternative training procedure that leverages existing samples from $\pi$ instead of access to the true score is provided in Appendix~\ref{app:dataset_resampling}.

The restraint width $s = \sigma_{t^*}/\alpha_{t^*}$ controls the mixing time
of the restrained Langevin sampler. The upper bound on $s$ comes
from the slowest degrees of freedom. If $p(\cdot\given z)$ spans a free-energy
barrier, restrained Langevin must cross it and mixes no faster than the
unrestrained chain. Fast dynamics like
bond vibrations and angle bending in molecular dynamics (MD) simulations in principle put a lower bound on $s$ through numerical stability, because a stiff restraint requires a small time step, but in practice this bound lies well below choices of interest.  The two bounds are independent, and their ratio
can span many orders of magnitude, giving wide latitude in the choice of $s$.
The parameter $s$ plays the same role as the force constant of the restraining potential in umbrella sampling simulations
but is determined by the noise schedule in the Gibbs sampling context rather than set by hand.

\begin{theorem}[Training correctness]
\label{thm:training}
Let $z\in\R^d$, let $x_0\sim p(\cdot\given z)$, and let $x_t$ be drawn from
the bridge \eqref{eq:bridge} with endpoint $y=\alpha_{t^*}z$.  Then for any
$\pi$, any $t^*\in(0,T)$, and every $t\in(0,t^*)$:
\begin{enumerate}
  \item[(i)] The $L^2$-optimal regressor of the score target
    $v = -(x_t-\alpha_t x_0)/\sigma_t^2$ against $x_t$ equals
    $\nabla_{x_t}\!\log \pi_t(x_t)$.
  \item[(ii)] The $L^2$-optimal regressor of the velocity target
    $u_t(x_t\given x_0) = \dot\alpha_t\,x_0 + \dfrac{\dot\sigma_t}{\sigma_t}(x_t - \alpha_t\,x_0)$
    against $x_t$ equals $f_t^\pi(x_t)$ in \eqref{eq:pf_ode}.
\end{enumerate}
\end{theorem}

\begin{proof}[Proof sketch]
Lemma~\ref{lem:posterior} identifies $x_0\sim\pi_{0\mid t^*}(\cdot\given y)$,
so the construction is a posterior-initialized likelihood bridge. The general
cancellation identity, Proposition~\ref{prop:likelihood_bridge}, gives part~(i),
and part~(ii) follows by an affine change of parameterization.
The full proof is in Appendix~\ref{app:likelihood_bridge}.
\end{proof}

Theorem~\ref{thm:training} extends immediately to a low-dimensional
collective variable (CV) $\mathcal{C}:\R^d\to\R^k$ constrained to a target
value $c\in\R^k$ (e.g.,\ a dihedral angle or end-to-end distance of a polymer). Define the  CV-restrained distribution to be  the joint conditional law
\begin{equation}
  \pi_{0,t^*\mid C_{t^*}}(x_0,x_{t^*}\given y)
  :=\mathcal L\!\left(x_0,x_{t^*}\,\middle|\,\mathcal C(x_{t^*})=c\right).
  \label{eq:joint_cv_target}
\end{equation}
For any draw
$(x_0,x_{t^*})$ from \eqref{eq:joint_cv_target}, the conditional $x_0\given x_{t^*}$
is exactly the probe-point restrained density \eqref{eq:restrained} with
implicit probe $z=x_{t^*}/\alpha_{t^*}$. The Gaussian bridge \eqref{eq:bridge} from $x_0$
to $x_{t^*}$ is therefore a standard probe-point bridge, and
Theorem~\ref{thm:training} applies directly for any $\mathcal{C}$ and $c$.
Appendix~\ref{app:cv_joint_sampling} gives the exact linear-CV construction.
For nonlinear CVs, the restrained data must be sampled directly from \eqref{eq:joint_cv_target}, rather than from a closed-form construction.

\begin{pseudocode}{Algorithm 1 \quad Restrained Bridge Training}
  \pcline{\pckw{Input:} pretrained $\psi_\theta$,\;
    bridge time $t^*$ (giving $\alpha_{t^*}$, $\sigma_{t^*}$),\;
    gradient steps $G$,\; batch size $B$,\; learning rate $\rho$}
  \pcline{\pckw{Input:} sample pools $\{\mathcal{D}_j\}_{j=1}^M$ providing pairs $(x_0,z_j)$:
    either $x_0\!\sim\! p(\cdot\given z_j)$ with explicit probe $z_j$ [probe-point],
    or $(x_0,x_{t^*})$ from \eqref{eq:joint_cv_target} with $z_j\!=\!x_{t^*}/\alpha_{t^*}$ [CV; Appendix~\ref{app:cv_joint_sampling}]}
  \pcline{\pckw{Output:} fine-tuned model $\psi_\theta$}
  \pcsep
  \pcline{\pckw{for} gradient step $g = 1$ \pckw{to} $G$ \pckw{do}}
  \pcline{\ind For each $b = 1,\ldots,B$: draw $j^{(b)} \sim \mathrm{Uniform}\{1,\ldots,M\}$
    and $(x_0^{(b)},\,z_{j^{(b)}})\sim \mathcal{D}_{j^{(b)}}$ independently
    \pccomment{one probe per item}}
  \pcline{\ind Sample bridge times $t^{(b)} \sim \mathrm{Uniform}(0,\, t^*)$ independently}
  \pcline{\ind \pckw{for} $b = 1$ \pckw{to} $B$ \pckw{do}}
  \pcline{\indd Compute bridge statistics for $(x_0^{(b)},\; z_{j^{(b)}},\; t^{(b)})$:}
  \pcline{\inddd $\mu^{(b)} \;\leftarrow\;
    \alpha_{t^{(b)}} x_0^{(b)}
    + \dfrac{\alpha_{t^*|t^{(b)}}\,\alpha_{t^*}\,\sigma_{t^{(b)}}^2}{\sigma_{t^*}^2}
    \!\bigl(z_{j^{(b)}} - x_0^{(b)}\bigr)$
    \pccomment{Eq.~\eqref{eq:mu}}}
  \pcline{\inddd $\sigma_{\mathrm{br}}^{(b)} \;\leftarrow\;
    \sigma_{t^{(b)}}\,\sigma_{t^*|t^{(b)}}\,/\,\sigma_{t^*}$
    \pccomment{Eq.~\eqref{eq:mu}}}
  \pcline{\indd Sample $\eta^{(b)} \sim \N(0,I)$;\quad
    set $x_t^{(b)} \leftarrow \mu^{(b)} + \sigma_{\mathrm{br}}^{(b)}\,\eta^{(b)}$}
  \pcline{\indd Compute denoising target $\tau^{(b)}$:
    \pccomment{Thm.~\ref{thm:training}; see below for both options}}
  \pcline{\inddd score: $\tau^{(b)} \leftarrow -(x_t^{(b)} - \alpha_{t^{(b)}} x_0^{(b)})/\sigma_{t^{(b)}}^2$}
  \pcline{\inddd velocity: $\tau^{(b)} \leftarrow \dot\alpha_{t^{(b)}} x_0^{(b)} + (\dot\sigma_{t^{(b)}}/\sigma_{t^{(b)}})(x_t^{(b)} - \alpha_{t^{(b)}} x_0^{(b)})$}
  \pcline{\ind \pckw{end for}}
  \pcline{\ind Update:\quad
    $\theta \leftarrow \theta - \rho\,\nabla_\theta\,\dfrac{1}{B}\displaystyle\sum_{b}
       \bigl\|\psi_\theta(x_t^{(b)},\, t^{(b)}) - \tau^{(b)}\bigr\|^2$}
  \pcline{\pckw{end for}}
  \pcline{\pckw{return} $\psi_\theta$}
\end{pseudocode}
The model $\psi_\theta$ need not represent every degree of freedom of the system.
It is sufficient that the restraint on the full system depend only on the modeled variables
(e.g.,\ protein backbone dihedral angles).
Let the full configuration $X \sim \Pi$ reduce to the modeled variable
$x_0 = R(X)$, with target marginal $\pi = R_{\#}\Pi$.
The CV conditioning in \eqref{eq:joint_cv_target} multiplies $\pi$ by a factor
$b_y(x_0)$ depending only on $x_0$, with
$p(x_0 \mid y) \propto \pi(x_0)b_y(x_0)$ the $x_0$-marginal of
\eqref{eq:joint_cv_target}.
Biasing the full system by $b_y(R(X))$ therefore produces this restrained marginal from training data generated in the full space.

Several strategies are available to select the probe points $z_j$. If an initial set of training samples from $\pi$ is available and contains one or more samples $x_\mathrm{tar}$ in the targeted region, then we can choose  $z_j = x_{\mathrm{tar}} + s\,\eta_j$ with $\eta_j\sim\N(0,I)$.   The points $x_\mathrm{tar}$ can be selected deterministically, or according to any chosen reweighting of the initial training samples from $\pi$. Note that each $x_\mathrm{tar}$ can correspond to many $z_j$. The added noise is not strictly necessary but is often important for training an accurate denoiser.

If few or no samples from $\pi$ are available, several exploration strategies can be used to generate probe points $x_\mathrm{tar}$ covering the target region. A simple option is to run Langevin dynamics on $\pi$ at reduced
inverse temperature $\beta < 1$ (scaling the score by $\beta$ samples
$\pi^\beta$). Other options include adiabatic-bias MD \citep{marchi1999adiabatic}, which can steer configurations to prescribed regions of sample space, as well as temperature-accelerated MD \citep{maragliano2006temperature}, and metadynamics \citep{Laio2002}.

Algorithm~1 handles multiple probe points by drawing one independent
$(z_{j^{(b)}},x_0^{(b)})$ pair per batch item, giving an unbiased estimator of
$\E_{z\sim q}\,\E_{x_0\sim p(\cdot|z)}[\nabla_\theta\ell_{\mathrm{bridge}}]$
where $\ell_{\mathrm{bridge}} = \|\psi_\theta(x_t, t) - \tau\|^2$ and $q$ is the distribution
of probe points.
The CV-restrained case (Appendix~\ref{app:cv_joint_sampling})
fits within the same loop. Probe-point and CV pools may be mixed across
$\{j\}$ with any chosen frequencies.
When $\nabla\log\pi$ is unavailable but  samples from $\pi$ exist,
dataset resampling (Appendix~\ref{app:dataset_resampling}) provides a third
pool type that fits the same loop without any modification.

\section{Inference}
\label{sec:inference}

\subsection{Inference from a fine-tuned model}
\label{sec:inference_finetune}

Theorem~\ref{thm:training} guarantees that the procedure in Algorithm~1 produces the correct
driving field of the reverse dynamics ($\nabla_{x_t}\!\log \pi_t$ or $f_t^\pi$) for $t\in [0,t^*]$.  Combined with the pretrained model on $[t^*,T]$, the resulting model on $[0,T]$ therefore produces samples correctly distributed according to $\pi$ (up to training error).  The $\pi_{t^*}$ marginal smooths $\pi$ by a Gaussian
kernel of variance $\sigma_{t^*}^2$, so the pretrained denoiser need only capture
the smoothed landscape. The fine-tuned model resolves local detail for
$t\le t^*$.  Errors arise only if the pretrained denoiser is inaccurate at noise
levels above $t^*$, because the reverse dynamics then deliver $x_{t^*}$ with a distribution that deviates from $\pi_{t^*}$.

\subsection{\texorpdfstring{Inference without $[t^*,T]$}{Inference without [t*, T]}}
\label{sec:gibbs}

When no pretrained denoiser is available for $t>t^*$, the bridge-trained model
supports a learned Gibbs sampler that produces correlated samples
on the joint density
\begin{equation}
  p(x_0, z) \;=\; \pi(x_0)\,\N\!\bigl(z;\, x_0,\, s^2 I\bigr),
  \label{eq:joint}
\end{equation}
whose $x_0$-marginal is $\pi$ by construction, and whose conditionals are
$p(x_0\given z)=\pi_{0\mid t^*}(x_0\given\alpha_{t^*}z)$ by Lemma~\ref{lem:posterior}
and $p(z\given x_0)=\N(z;x_0,s^2I)$.
Gibbs sampling on \eqref{eq:joint} alternates between two steps:
\begin{enumerate}
  \item \textbf{Sample} $x_0\sim\pi_{0\mid t^*}(\cdot\given\alpha_{t^*}z)$: set
    $x_{t^*}\leftarrow\alpha_{t^*}z$ exactly and run the stochastic reverse SDE
    \eqref{eq:reverse_sde} from $t^*$ to $0$ with $\psi_\theta$.
  \item \textbf{Resample} $z\sim\N(x_0,s^2 I)$: a single Gaussian draw.
\end{enumerate}
The $z$-update ($z\leftarrow x_0+s\eta$, $\eta\sim\N(0,I)$) samples exactly from the forward kernel
$x_{t^*}\given x_0\sim\N(\alpha_{t^*}x_0,\sigma_{t^*}^2 I)$, so the resulting $x_{t^*}$
has the correct marginal $\pi_{t^*}$.
Stochastic reverse diffusion from $x_{t^*}$ with the correct score
(Theorem~\ref{thm:training}) then
recovers $\pi_{0\mid t^*}(x_0\given x_{t^*})$,
and marginalizing over $x_{t^*}$ returns $\pi$.
Pseudocode, proximal interpretation, and a Metropolized variant are  provided in Appendix~\ref{app:ncmc}.

\section{Numerical Experiments}
\label{sec:experiments}

We evaluate restrained diffusion bridge (RDB) refinement on the two-dimensional M\"uller--Brown system, as well as two molecular generation problems:
isomerization of \textit{N}-acetyl-\textit{N}$^\prime$-methylalaninamide, often termed the the alanine dipeptide since it is an alanine amino acid with two peptide bonds to blocking group, in explicit solvent demonstrates Cartesian sampling
without an explicit solute-marginal energy function, while
folding of the 10-amino-acid peptide chignolin in implicit solvent tests high-dimensional rare-state fine-tuning in the latent space of a
model with 93 heavy (i.e., nonhydrogen) atoms. We assess both rare-state and global accuracy.

\subsection{M\"uller--Brown potential}
\label{sec:mueller_brown_exp}
The M\"uller--Brown (MB) potential $U(x)$ \citep{Muller1979} is a two-dimensional energy surface with three minima (metastable basins). The Boltzmann density
$\pi(x) \propto \exp(-\beta U(x))$ can be evaluated exactly on a grid. For this example we measure sampling quality
against results from quadrature. The inverse-temperature $\beta$ acts as a difficulty
parameter. Raising it sharpens the gradients of the target and raises the effective
barriers between minima. In this section, we examine the performance of restrained RDB training with a fixed budget for generating the training data and Gibbs sampling inference in $\beta\in \{1,1.5,2,3\}$. Our goal is to understand how model accuracy changes as the problem becomes more difficult. For the MB example we do not consider the cost of inference. More details can be found in
Appendix~\ref{app:mb-tstar}.

We train a  model with RDB on restrained Langevin bridge
endpoints alone, without pretraining, over diffusion times $t \in [0, t^\ast]$ for $t^\ast = 0.02$. This RDB model is sampled at inference time using the Gibbs strategy described in Section~\ref{sec:gibbs}. We obtain $10{,}000$ probe points $z$ from an unrestrained Langevin trajectory at $\beta_{\mathrm{probe}} = 0.5$, recorded every
$0.25$ time units. Each probe initializes a 0.3 time-unit trajectory of overdamped
Langevin dynamics, whose final configuration yields a single restrained sample. Integrations were performed at a time step of $\mathrm{d}t = 0.001$.
We train the baseline standard diffusion model over the full range $t \in [0, 1]$ on unrestrained Langevin data at each $\beta$ value considered. The two models use the same architecture (a three-layer multilayer perceptron (MLP) with
$128$ hidden units, trained for $10{,}000$ iterations) and total Langevin   cost, which yields $10{,}000$ training points for the RDB trained model and $22{,}000$ points for the standard diffusion model at each $\beta$ value.

For the RDB trained model with Gibbs sampling inference on $[0,t^\ast]$, we run $1{,}000$ independent inference chains, all initialized at the deepest minimum, for $50{,}000$ sweeps each. Each sweep noises the state forward to $t^\ast$ and integrates the reverse process back to $t=0$ in $100$ steps. We treat sweeps before $10{,}000$ as burn-in and retain every tenth configuration from sweep $10{,}000$ through sweep $50{,}000$, inclusive. This yields $4{,}001$ configurations per chain and $4{,}001{,}000$ pooled samples. The baseline is sampled directly from noise using the same $100$ reverse steps, yielding $4{,}000{,}000$ samples.

\begin{figure}[t]
\centering
\includegraphics[width=\textwidth]{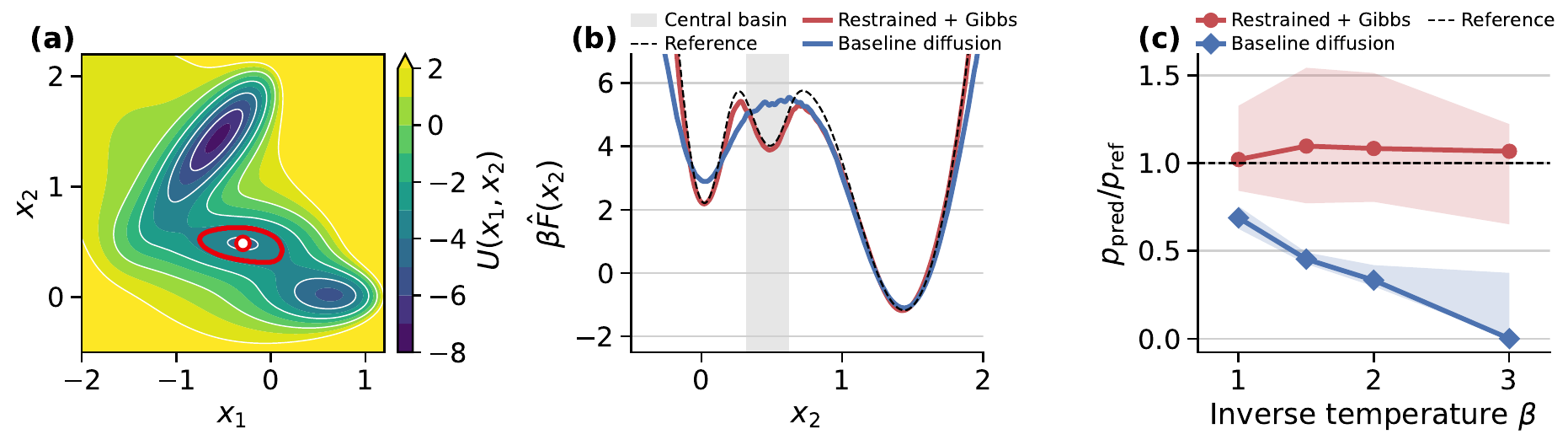}
\caption{
    (a) The M\"uller--Brown potential with the central basin outlined in red.
    (b) Free-energy profile along $x_2$, $\beta \hat{F}(x_2) = -\ln p(x_2)$, at $\beta = 2.00$ for a representative sampling run of each method against the exact profile; the shaded band marks the extent
    of the central basin in $x_2$. The restrained sampler recovers the central basin, whereas the baseline does not.
    (c) Central-basin occupancy relative to its reference value from quadrature,
    $p_{\mathrm{pred}} / p_{\mathrm{ref}}$, against inverse temperature $\beta$ for the restrained
    sampler at $t^\ast = 0.02$ and for the budget-matched baseline. Raw occupancy values are plotted in Appendix~\ref{app:mb-tstar}. Median (dots and diamonds) and interquartile range (shading) are plotted for twenty independently seeded inference runs.
}
\label{fig:mb_results}
\end{figure}

The central basin in Figure~\ref{fig:mb_results}(a) is high in energy compared with the other two and carries little
probability mass. A
diffusion model trained directly on equilibrium samples therefore sees relatively few training points there.
As shown in Figure~\ref{fig:mb_results}(b), the baseline fails to resolve this basin,
whereas the RDB trained model with Gibbs sampling inference captures it accurately.
Figure~\ref{fig:mb_results}(c) shows that the failure of the baseline model becomes more acute as temperature decreases, with its interquartile range excluding exact agreement at every
$\beta$. In contrast, the restrained sampler stays close to the exact occupancy throughout the temperature range, suggesting that an analogue of umbrella sampling's favorable low-temperature efficiency \citep{Dinner2020} may apply.

\subsection{Alanine dipeptide in explicit solvent}

\begin{figure}[t]
\centering
\includegraphics[width=\textwidth]{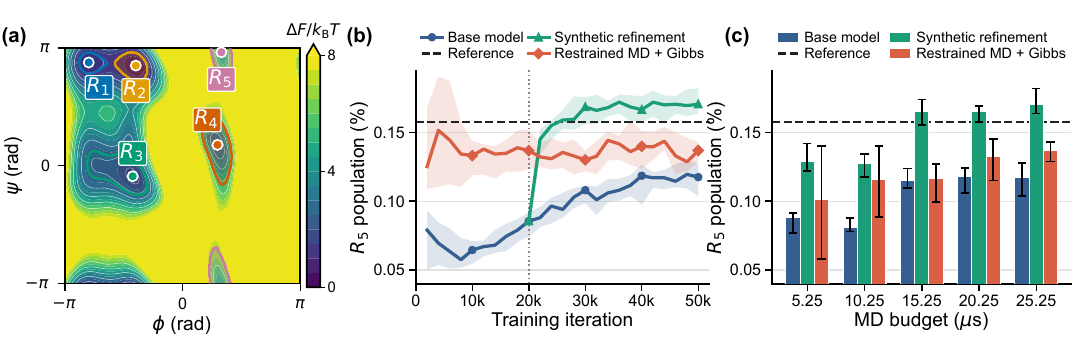}
\caption{
Results for isomerization of the alanine dipeptide in explicit solvent. (a) Free energy as a function of the Ramachandran
angles, with the five basins that we characterize marked. (b) Generated \(R_5\) population versus training updates at a fixed
\(25.25\,\mu\mathrm{s}\) MD budget; synthetic-pair refinement branches at 20{,}000 updates.
(c) Generated \(R_5\) population versus matched MD budget at 50{,}000 updates.
Curves and bars show medians over 20 training runs, and bands and error bars show
interquartile ranges. Dashed lines indicate the reference value.
}
\label{fig:NANMA_results}
\end{figure}

Our alanine dipeptide numerical experiment involves a 22-atom solute and 628 TIP3P water molecules~\citep{tip3pwater}.
Solvent variables appear only in the MD simulations used to generate training data.
The diffusion model generates (Kabsch-aligned~\citep{kabsch}) Cartesian coordinates for the solute atoms. We use Ramachandran angles \((\phi,\psi)\)~\citep{Ramachandran} to  define five metastable free-energy basins, including the
rarest, \(R_5\) (Figure~\ref{fig:NANMA_results}(a)). We assess global fidelity using the
Jensen--Shannon (JS) distance between normalized two-dimensional
\((\phi,\psi)\) histograms, defined as the square root of the JS
divergence~\citep{lin1991divergence}. Rare-state accuracy is assessed by
comparing the generated population \(p_{\mathrm{gen}}(R_5)\) with the reference population.

For both the reference population and training the baseline diffusion model, we run five 300 K equilibrium MD trajectories of \(5.05\,\mu\mathrm{s}\) each, including a \(50\,\mathrm{ns}\) warm-up per trajectory, for a total MD budget of \(25.25\,\mu\mathrm{s}\). The baseline model captures the global distribution but underestimates the population of \(R_5\).

We compare the baseline diffusion model to two ways to improve rare-basin sampling.
The two approaches use the same model architecture and are compared at matched numbers of training updates.

\paragraph{Setting I: Refinement without additional MD.}
We use the equilibrium dataset to construct synthetic restraint pairs and fine-tune the baseline model from its \(20{,}000\)-update checkpoint for another \(30{,}000\) updates. On a \(50\times50\) grid on $(\phi, \psi)$, \(10{,}000\) probes are allocated in proportion to \(\sqrt{n_{ij}}\), where \(n_{ij}\) is the equilibrium count in cell \((i,j)\). For each probe, we select 30 synthetic restrained samples without further MD or energy evaluations (Appendix~\ref{app:dataset_resampling}). The restraints act on the Cartesian coordinates of the solute heavy atoms (Appendix~\ref{app:cv_joint_sampling}); the Ramachandran angles are used only to allocate probes and evaluate the generated samples.
Following refinement, inference consists of using the pretrained model on $[t^*,T]$ and the fine-tuned model on $[0,t^*]$.

\paragraph{Setting II: Generation at fixed MD cost.}
When starting from an energy function and no pre-trained model, we instead spend the same \(25.25\,\mu\mathrm{s}\) MD budget on a \(250\,\mathrm{ns}\) trajectory at 500 K to generate $5{,}000$ probes  and \(5{,}000\times5\,\mathrm{ns}\) of restrained MD at 300 K. Although total simulation time is matched, the restrained trajectories allow much greater parallelism and can substantially reduce wall-clock time. The resulting pairs train a denoiser on \([0,t^*]\), with \(t^*=0.40\), from scratch. Equilibrium samples are then obtained using a Gibbs chain.

At the full \(25.25\,\mu\mathrm{s}\) MD budget, Figure~\ref{fig:NANMA_results}(b) compares two variants of RDB refinement. The synthetic-pair variant branches from the baseline model at 20{,}000 updates and rapidly brings the \(R_5\) population close to the reference. The restrained-MD-pair variant provides a more accurate estimate than the baseline model even at the earliest training checkpoints. To assess sensitivity to simulation cost, we then fix training at \(50{,}000\) updates and vary the matched MD budget from \(5.25\) to \(25.25\,\mu\mathrm{s}\) (Figure~\ref{fig:NANMA_results}(c)). At each budget, we shorten all five equilibrium trajectories used by the baseline model and synthetic-pair refinement, and adjust the number of restrained-MD probes from \(1{,}000\) to \(5{,}000\). Synthetic-pair refinement improves the \(R_5\) estimate at all budgets. Restrained-MD refinement yields a more accurate estimate than the matched-budget baseline model and continues to improve with increasing MD budget, whereas the baseline model accuracy appears to saturate. The 2D JS distances remain similar across methods and budgets. Their trends and further details are reported in Appendix~\ref{app:NANMA}.

\subsection{Chignolin in implicit solvent}

\begin{figure}[t]
\centering
\includegraphics[width=\textwidth]{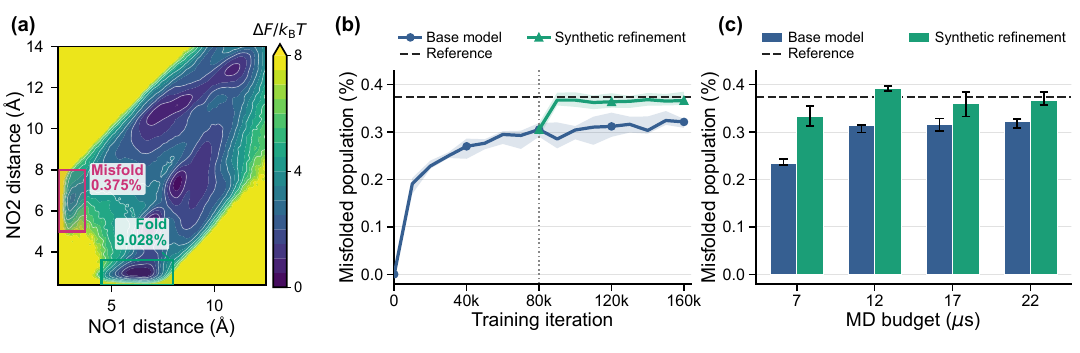}
\caption{Results for chignolin in implicit solvent. (a) Free energy in the
$(\mathrm{NO1},\mathrm{NO2})$ plane estimated from the $20\,\mu\mathrm{s}$ reference
ensemble, with folded and misfolded regions outlined. (b) Generated
misfolded population versus training updates at a fixed $22\,\mu\mathrm{s}$
MD budget. Synthetic-pair refinement branches at 80{,}000 updates.
(c) Generated misfolded population versus total MD budget
at 160{,}000 total updates. Synthetic-pair refinement uses $t^*=0.30$. Curves and bars
show medians over 10 training runs, and bands and error bars show interquartile ranges.
Dashed lines indicate the reference value.
}
\label{fig:chignolin}
\end{figure}

Next we study a variant of the peptide chignolin (CLN025, with sequence Tyr-Tyr-Asp-Pro-Glu-Thr-Gly-Thr-Trp-Tyr, corresponding to 93 heavy atoms)~\citep{chig}
using a $22\,\mu\mathrm{s}$ trajectory at $340\,\mathrm{K}$, sampled
every $10\,\mathrm{ps}$. We discard the first $2\,\mu\mathrm{s}$ as warm-up,
and evaluate every model against the remaining $20\,\mu\mathrm{s}$
reference ensemble.  The distances
$\mathrm{NO1}=d(\mathrm{Asp3\,N},\mathrm{Gly7\,O})$ and
$\mathrm{NO2}=d(\mathrm{Asp3\,N},\mathrm{Thr8\,O})$ distinguish the native
$\pi$-turn (short NO2) from the misfolded $\alpha$-turn (short NO1,
longer NO2)~\citep{turn}.  The misfolded region,
$\mathrm{NO1}<3.7\,\text{\AA}$ and
$5.0<\mathrm{NO2}<8.0\,\text{\AA}$, has reference population
$0.3747\%$ (Figure~\ref{fig:chignolin}(a)).

For this system, we did not obtain a satisfactory Cartesian
baseline diffusion model within a reasonable training budget. We therefore represent the configuration with 128 latent variables of an autoencoder trained on the MD data for each MD budget. We use the same trained autoencoder across independently seeded diffusion runs at each MD budget (Appendix~\ref{app:chig}).  Restraining to the latent variables would require repeated encoder evaluation and differentiation during MD simulation, so we
consider only refinement without additional MD simulation (Setting~I).  As for the alanine dipeptide, we
allocate $10{,}000$ probes to occupied $(\mathrm{NO1},\mathrm{NO2})$ cells
in proportion to $\sqrt{n_{ij}}$ and construct 30 synthetic posterior
endpoints per probe.  The baseline model is trained for $160{,}000$ updates. Synthetic-pair refinement branches from its 80{,}000 update checkpoint and performs 80{,}000
fine-tuning updates at $t^*$. Following refinemet, inference consists of  using the baseline model on $[t^*,T]$ and the fine-tuned model on $[0,t^*]$.

At the full $22\,\mu\mathrm{s}$ MD budget, the median generated
misfolded population rises from $0.3055\%$ at the 80{,}000 update branch point to
$0.3675\%$ after only 10{,}000 refinement updates and remains close to the
reference thereafter (Figure~\ref{fig:chignolin}(b)).  At 160{,}000 total
updates, synthetic-pair refinement improves the median from $0.3215\%$ for the
baseline model to $0.3665\%$, reducing the absolute relative error from
$14.2\%$ to $2.2\%$.  Across the $7$, $12$, $17$, and
$22\,\mu\mathrm{s}$ MD budgets, the corresponding synthetic
errors are $11.4\%$, $4.9\%$, $3.9\%$, and $2.2\%$, compared with
$37.8\%$, $16.5\%$, $15.7\%$, and $14.2\%$ for the baseline model
(Figure~\ref{fig:chignolin}(c)).
Additional JS-distance
and bridge-time results are reported in Appendix~\ref{app:chig}.

\section{Conclusions}
\label{sec:conclusion}

Leveraging the fact that the restrained density at scale $s$ is the exact diffusion posterior at the noise level $t^*$ matched to $s$ (Lemma~\ref{lem:posterior}), we showed that conditioning the forward noising process on restrained data exactly cancels the biasing effect of those data. As a result, a model trained on restrained distributions, using standard denoising or flow matching objectives, recovers the unconditional score of $\pi$ without importance weights or corrections (Theorem~\ref{thm:training}).

Algorithm~1 uses restrained Langevin samples to fine-tune a pretrained model on $[0,t^*]$ near chosen probe points. The result produces independent samples from $\pi$ via standard reverse dynamics at no extra cost.
A likelihood-free variant replaces Langevin sampling with importance-weighted resampling from an existing dataset when $\nabla\log\pi$ is unavailable (Appendix~\ref{app:dataset_resampling}).
When no pretrained model is available, the learned Gibbs sampler can use only the bridge-trained model on $[0,t^*]$ to generate correlated samples from $\pi$.

Our approach cannot repair errors due to an inadequate choice of network architecture. We are also not yet able to provide more than heuristic guidance on the choice of $t^*$, though we find that our results are not overly sensitive to that choice (Appendix~\ref{sec:metrics}). For the Gibbs scheme there is an important tradeoff between higher sampling efficiency of the restrained distributions for lower $t^*$ and faster mixing at inference for larger $t^*$ (Appendix~\ref{app:mb-tstar}).

Molecular simulation, where large pretrained structure models degrade at transition states and rare configurations that often govern biological function, is a natural application. Similar problems arise in other fields such as generative models for global climate and image generation. Algorithm 1 addresses this failure without global retraining.

Several directions remain open: whether an analog of the umbrella sampling exponential data efficiency result in \citet{Dinner2020} holds, extensions of the framework to discrete diffusion, and broader applications of the likelihood-free resampling approach to high-dimensional generative settings such as image generation.

\subsection*{Acknowledgments}

 We acknowledge support from National Institutes of Health awards R35~GM136381 and R01~GM169399, Office of Naval Research award N000142612500, and the Margot and Tom Pritzker Science Foundation.
 Generative AI tools were used to identify relevant literature, create and revise scientific figures, and write and revise code used in the execution of the research. Generative AI tools were also used to draft and revise portions of the manuscript and to improve its clarity, structure, grammar, and style. The authors take responsibility for the final content of this paper, including text, claims, and artifacts produced with the assistance of generative AI.

\bibliographystyle{plainnat}
\bibliography{refs}
\appendix

\section{Likelihood--bridge cancellation and proof of Theorem~\ref{thm:training}}
\label{app:likelihood_bridge}

The main theorem is a Gaussian instance of the following classical
Doob-transform identity.
\begin{proposition}[Posterior-initialized likelihood bridge]
\label{prop:likelihood_bridge}
Let a Markov process $x_t$ have initial density $\pi$, transition density
$k_{0,t}(x\mid x_0)$, and a nonnegative endpoint-observation likelihood
$\ell_y$. For $t\le t^*$ define
\[
  h_t(x)=\E[\ell_y(x_{t^*})\mid x_t=x].
\]
Initialize $x_0$ from the posterior proportional to $\pi(x_0)h_0(x_0)$
and sample the corresponding likelihood-conditioned bridge. Then the
population regressor of the ordinary denoising target
$\nabla_x\log k_{0,t}(x\mid x_0)$, conditional on $x_t=x$, is
$\nabla_x\log \pi_t(x)$.
\end{proposition}
\begin{proof}
Writing $Z_y=\int\pi(x_0)h_0(x_0)\,dx_0$, the posterior-initialized bridge
joint density at times $0$ and $t$ is
\[
  q_y(x_0,x)=Z_y^{-1}\pi(x_0)k_{0,t}(x\mid x_0)h_t(x).
\]
The factor $h_t(x)$ is constant in the conditional law of $x_0$ given
$x_t=x$. Hence
\[
 \E_{q_y}[\nabla_x\log k_{0,t}(x\mid x_0)\mid x_t=x]
 =\nabla_x\log\!\int \pi(x_0)k_{0,t}(x\mid x_0)\,dx_0
 =\nabla_x\log \pi_t(x).
\]
Thus the likelihood, equivalently the Doob-transform factor, cancels from the
ordinary regression.
\end{proof}

\begin{proof}[Proof of Theorem~\ref{thm:training}]
For the Gaussian process, observing $x_{t^*}=y$ gives
$h_t(x)=k_{t,t^*}(y\mid x)$. Lemma~\ref{lem:posterior} identifies the
posterior initial law $\pi(x_0)h_0(x_0)/Z_y$ with $p(x_0\given z)$, and the
corresponding likelihood-conditioned process is the bridge \eqref{eq:bridge}.
Moreover,
\[
  \nabla_{x_t}\log k_{0,t}(x_t\mid x_0)
  =-\frac{x_t-\alpha_t x_0}{\sigma_t^2}=v.
\]
Proposition~\ref{prop:likelihood_bridge} proves part~(i). For part~(ii),
$g_t^2=\dot\sigma_t^2-2(\dot\alpha_t/\alpha_t)\sigma_t^2$ gives
\[
  u_t(x_t\given x_0)
  =\frac{\dot\alpha_t}{\alpha_t}x_t-\frac{g_t^2}{2}v.
\]
Taking conditional expectations and applying part~(i) yields
$\E[u_t\mid x_t]=f_t^\pi(x_t)$.
\end{proof}

The exact endpoint observation gives the Gaussian restrained construction,
while the regular conditional observation $\mathcal C(x_{t^*})=c$ gives
\eqref{eq:joint_cv_target}. Thus both the probe-point and exact CV constructions
are special cases of Proposition~\ref{prop:likelihood_bridge}, which is an
organizing identity rather than a novelty claim about the Doob transform.

\section{Sampling the CV-restrained joint density}
\label{app:cv_joint_sampling}

This appendix provides procedures for generating pairs
$(x_0^{(i)},x_{t^*}^{(i)})$ from the conditional law \eqref{eq:joint_cv_target},
which are passed to Algorithm~1 with implicit probe
$z_j = x_{t^*}^{(i)}/\alpha_{t^*}$.

\paragraph{Linear CVs.}
Let $\mathcal{C}(x)=Ax$ with $A\in\R^{k\times d}$ full row-rank.
Marginalizing $x_{t^*}$ in \eqref{eq:joint_cv_target} gives the
$x_0$-marginal
\begin{equation}
  \pi_{0\mid C_{t^*}}(x_0 \given y)
  \;\propto\;
  \pi(x_0)\,\N\!\bigl(y;\,\alpha_{t^*}Ax_0,\,\sigma_{t^*}^2 AA^\top\bigr),
  \label{eq:cv_x0_marginal}
\end{equation}
a harmonic restraint on $Ax_0$ with target $y/\alpha_{t^*}$ and stiffness
$\alpha^2_{t^*}/\sigma^2_{t^*}$. Samples $x_0^{(i)}$ are drawn by restrained Langevin
dynamics in $\R^d$.  Given $x_0^{(i)}$, the $x_{t^*}$-conditional is
Gaussian and is drawn analytically:
\begin{equation}
  x_{t^*}^{(i)} \sim \N\!\Bigl(
    \alpha_{t^*}x_0^{(i)} + A^\top(AA^\top)^{-1}
    \bigl(y - \alpha_{t^*}Ax_0^{(i)}\bigr),\;
    \sigma_{t^*}^2\bigl(I - A^\top(AA^\top)^{-1}A\bigr)
  \Bigr).
  \label{eq:cv_xt_conditional}
\end{equation}
Since $I-A^\top(AA^\top)^{-1}A$ is the orthogonal projector onto $\ker A$,
the noise lies in $\ker A$ and $Ax_{t^*}^{(i)}=y$ holds exactly.
Pairs $(x_0^{(i)},x_{t^*}^{(i)})$ are therefore generated independently
with no MCMC over $x_{t^*}$.

\paragraph{Nonlinear CVs.}
When $\mathcal{C}$ is nonlinear, the sampler must target the regular
conditional law \eqref{eq:joint_cv_target}. Replacing the constraint by a
finite stiff harmonic penalty gives  an approximation to that law that can be sampled using standard algorithms.

Setting $\mathcal{C}(x)=x$ (i.e.\ $A=I$, $k=d$) in \eqref{eq:cv_xt_conditional},
the covariance $\sigma_{t^*}^2(I - I^\top(II^\top)^{-1}I)$ vanishes, so
the conditional distribution degenerates to the point mass $x_{t^*}=y=\alpha_{t^*}z$.
The two-step procedure then reduces exactly to the probe-point construction
of Section~\ref{sec:algorithms}: sample $x_0\sim p(\cdot\given z)$ by
restrained Langevin, then set $x_{t^*}=\alpha_{t^*}z$ deterministically.

\section{Dataset resampling for likelihood-free bridge-based model refinement}
\label{app:dataset_resampling}

Algorithm~1 requires $\nabla\log\pi$ to draw restrained samples by Langevin
dynamics.
When   samples $\mathcal{D}=\{x_0^{(i)}\}_{i=1}^m$ from $\pi$ are available but gradients are not---e.g.,\
image distributions or black-box simulators---restrained samples can instead
be obtained by importance-weighted resampling from
$\mathcal{D}$.

Choose a probe by perturbing a dataset point: $z=\tilde{x}+s\eta$,
$\tilde{x}\in\mathcal{D}$, $\eta\sim\N(0,I)$.
Assign each dataset point the weight
\begin{equation}
  w^{(i)}(z) \;=\; \exp\!\Bigl(-\frac{\norm{x_0^{(i)}-z}^2}{2s^2}\Bigr)
  \label{eq:iw_weights}
\end{equation}
and draw $x_0^*$ proportional to the normalized weights
$\bar{w}^{(i)}(z)=w^{(i)}(z)/\sum_j w^{(j)}(z)$.
The pair $(x_0^*,z)$ replaces the Langevin-drawn pair in Algorithm~1.
All other steps are unchanged.

\begin{proposition}[Dataset-resampling correctness]
\label{prop:dataset_resampling}
Let $x_0^*$ be drawn with weights~\eqref{eq:iw_weights}.
As $m\to\infty$, $x_0^*$ converges weakly to $p(\cdot\given z)$, and the
$L^2$-minimizer of the bridge training loss converges to
$\nabla_{x_t}\!\log \pi_t$ (score parameterization) or $f_t^\pi$ (velocity
parameterization).
\end{proposition}
\begin{proof}
Let $w_z(x)=\exp(-\|x-z\|^2/2s^2)$. For any bounded continuous $\varphi$,
the self-normalized law of large numbers gives, under standard integrability
conditions,
\[
 \sum_i\bar w^{(i)}(z)\varphi(x_0^{(i)})
 \longrightarrow
 \frac{\E_\pi[w_z(x)\varphi(x)]}{\E_\pi[w_z(x)]}
 =\E_{p(\cdot\given z)}[\varphi(x)].
\]
Thus the empirical weights are proportional only to $w_z(x_i)$, while the
factor $\pi$ is supplied by drawing the dataset from $\pi$.
The result then follows by Theorem~\ref{thm:training}.
\end{proof}

The weights in \eqref{eq:iw_weights} are large only for dataset points within
distance $\sim s$ of $z$. Points farther away contribute negligibly.
The method therefore fails when $z$ falls in a low-density region of $\pi$,
where few dataset points lie nearby and the resampled $x_0^*$ is determined by
whichever point happens to be closest rather than by the true conditional
$p(\cdot\given z)$.
Restrained Langevin handles this case because it actively drives $x_0$ toward
$z$. Resampling is the better choice when the dataset already covers the
neighborhood of $z$ but the pretrained model is miscalibrated there.

\section{Gibbs sampling details}
\label{app:ncmc}

\begin{pseudocode}{Algorithm 2 \quad Generative Gibbs Sampler}
  \pcline{\pckw{Input:} fine-tuned $\psi_\theta$,\;
    $\alpha_{t^*}$,\; $s = \sigma_{t^*}/\alpha_{t^*}$,\;
    initial probe $z_0$,\; iterations $K$,\; thinning $\kappa$}
  \pcline{\pckw{Output:} samples $\{x_0^{(k)}\}_{k=1}^K$ from $\pi$ (exactly when $\psi_\theta$ is exact)}
  \pcsep
  \pcline{Initialize $z \leftarrow z_0$}
  \pcline{\pckw{for} $k = 1$ \pckw{to} $K\kappa$ \pckw{do}}
  \pcline{\ind Set $x_{t^*} \leftarrow \alpha_{t^*}\,z$
    \pccomment{exact bridge endpoint; bypasses $t>t^*$ entirely}}
  \pcline{\ind Run the stochastic reverse SDE \eqref{eq:reverse_sde} from $t^*$ to $0$ with $\psi_\theta$
    to obtain $x_0$
    \pccomment{Thm.~\ref{thm:training}}}
  \pcline{\ind Sample $\eta\sim\N(0,I)$;\quad set $z \leftarrow x_0 + s\,\eta$
    \pccomment{$z\given x_0 \sim \N(x_0, s^2 I)$, no score evaluation}}
  \pcline{\ind \pckw{if} $\kappa\mid k$ \pckw{then}
    record $x_0^{(k/\kappa)} \leftarrow x_0$ \pckw{end if}}
  \pcline{\pckw{end for}}
  \pcline{\pckw{return} $\{x_0^{(k)}\}$}
\end{pseudocode}

Algorithm~2 is a learned proximal Gibbs sampler: its first step samples the
restricted-Gaussian conditional with stochastic reverse diffusion, while its
second step is an exact Gaussian update \citep{Lee2021,Kim2023,Chen2024}.
Step~1 is called a \emph{proximal sample} of $\pi$ at $z$ because
$p(\cdot\given z)\propto\pi\exp(-\|\cdot-z\|^2/2s^2)$ is the stochastic analog
of the proximal operator familiar from convex optimization,
$\mathrm{prox}_{f,s}(z)=\arg\min_x\{f(x)+\tfrac{1}{2s^2}\|x-z\|^2\}$,
with $f = -\log\pi$.  Rather than minimizing a penalized objective, we
sample from the corresponding penalized distribution. With Metropolization, the construction here is essentially identical to that introduced recently in \citet{chen2026markov}. The key distinction is that here we use restrained data to train the sampler, avoiding the need for $\pi$ or $\nabla \log \pi$ during training and permitting direct training of marginal samplers.

Algorithm~2 targets $\pi$ exactly when $\psi_\theta$ is exact
(Theorem~\ref{thm:training}), and in practice the error is entirely
attributable to the quality of the fine-tuned model and time discretization error, not to the Gibbs
structure.  Mixing speed depends on how broadly the probe points $\{z_j\}$
cover the relevant region during training and on the choice of $s$. Matching
$s$ to the local length scale of $\pi$ improves both coverage and mixing.
Algorithm~2 produces correlated rather than independent samples, but because
$t^*$ is small relative to $T$, each Step~1 requires only a fraction of the
function evaluations of a full diffusion trajectory.

Though we do not employ it in any of our experiments in this paper,
when model error is a concern, Metropolization via Nonequilibrium Candidate Monte
Carlo (NCMC) \citep{Nilmeier2011} is straightforward and eliminates asymptotic bias entirely.
The Metropolized chain then targets $\pi$ exactly, regardless of model quality, at the cost of
occasional rejected proposals.  As with all Metropolized samplers, the rejection rate will increase with dimension, even when the un-Metroplized sampler's accuracy does not necessarily degrade rapidly~\citep{chen2026deloc}.

Metropolization by NCMC requires $\psi_\theta$ to use a
score parameterization ($\psi_\theta\approx\nabla_{x_t}\!\log \pi_t$),
written $s_\theta$ in what follows.  We write
$p_\theta(\cdot\given x_{t_n})$ for the one-step reverse-transition density
obtained by an Euler--Maruyama step of the reverse SDE~\eqref{eq:reverse_sde}
with score $s_\theta$, and $\overleftarrow{X}$ for the time-reversal of a
discrete trajectory (i.e.\ the sequence with its elements in reversed order).
Computing the path log-probabilities $\log p_\theta$ requires the SDE drift,
which depends explicitly on the score.

To construct the Metropolized scheme, we replace the exact marginal noising step
($z \leftarrow x_0 + s\eta$, $\eta\sim\N(0,I)$) of Algorithm~2 with a
discrete simulation of the forward SDE over the time grid
$0 = t_0 < t_1 < \cdots < t_N = t^*$, with Gaussian transition kernel
\[
  q(x_{t_n}\given x_{t_{n-1}})
  \;=\; \N\!\bigl(x_{t_n};\,\alpha_{t_n|t_{n-1}}\,x_{t_{n-1}},\,
    \sigma_{t_n|t_{n-1}}^2 I\bigr).
\]
This replaces the explicit probe anchor $z$ with a stochastic coupling state
$x_{t^*}$ whose path probability is tractable.  Denote the forward trajectory
$X_{\mathrm{fwd}}=(x_0,x_{t_1},\dots,x_{t^*})$ and the reverse denoising
trajectory $X_{\mathrm{rev}}=(x'_{t^*},x'_{t_{N-1}},\dots,x'_0)$ produced by
running $s_\theta$ via Euler--Maruyama from the shared coupling
state $x'_{t^*}=x_{t^*}$.  The full round-trip from $x_0$ to $x'_0$
constitutes a single NCMC proposal with acceptance probability
$A=\min(1,\exp(-W))$, where the nonequilibrium work is
\begin{align}
  W \;=\;
  &\underbrace{\bigl[\log\pi(x_0)-\log\pi(x'_0)\bigr]}_{W_{\mathrm{energy}}}
  \;+\;
  \underbrace{\bigl[\log q(X_{\mathrm{fwd}}\given x_0)
    -\log p_\theta(\overleftarrow{X_{\mathrm{fwd}}}\given x_{t^*})
  \bigr]}_{W_{\mathrm{out}}}
  \nonumber\\[4pt]
  &+\;
  \underbrace{\bigl[\log p_\theta(X_{\mathrm{rev}}\given x_{t^*})
    -\log q(\overleftarrow{X_{\mathrm{rev}}}\given x'_0)
  \bigr]}_{W_{\mathrm{in}}},
  \label{eq:ncmc_work}
\end{align}

\bigskip
\begin{pseudocode}{Algorithm 3 \quad Metropolized Generative Gibbs Sampling (NCMC-Gibbs)}
\pcline{\pckw{Input:} fine-tuned $s_\theta$ (score parameterization),\; true log-target
  $\log\pi$,\; time grid $0{=}t_0{<}t_1{<}\dots{<}t_N{=}t^*$,\;
  initial $x_0$,\; iterations $K$,\; thinning $\kappa$}
\pcline{\pckw{Output:} samples $\{x_0^{(k)}\}_{k=1}^K$ targeting $\pi$ exactly}
\pcsep
\pcline{\pckw{for} $k = 1$ \pckw{to} $K\kappa$ \pckw{do}}
\pcline{\ind \pccomment{1. Forward noising (outward leg)}}
\pcline{\ind Initialize $X_{\mathrm{fwd}} \leftarrow (x_0)$}
\pcline{\ind \pckw{for} $n = 1$ \pckw{to} $N$ \pckw{do}}
\pcline{\indd Sample $x_{t_n} \sim q(\cdot\given x_{t_{n-1}})$
  \pccomment{analytic Gaussian kernel}}
\pcline{\indd Append $x_{t_n}$ to $X_{\mathrm{fwd}}$}
\pcline{\ind \pckw{end for};\quad set coupling state $x_{t^*} \leftarrow x_{t_N}$}
\pcline{\ind \pccomment{2. Reverse denoising (inward leg)}}
\pcline{\ind Initialize $X_{\mathrm{rev}} \leftarrow (x_{t^*})$;\quad
  set $x'_{t_N} \leftarrow x_{t^*}$}
\pcline{\ind \pckw{for} $n = N$ \pckw{downto} $1$ \pckw{do}}
\pcline{\indd Sample $x'_{t_{n-1}} \sim p_\theta(\cdot\given x'_{t_n})$
  \pccomment{Euler--Maruyama with $s_\theta$}}
\pcline{\indd Append $x'_{t_{n-1}}$ to $X_{\mathrm{rev}}$}
\pcline{\ind \pckw{end for};\quad set candidate $x'_0 \leftarrow x'_{t_0}$}
\pcline{\ind \pccomment{3. Nonequilibrium work (Eq.~\eqref{eq:ncmc_work})}}
\pcline{\ind $W_{\mathrm{energy}} \leftarrow \log\pi(x_0) - \log\pi(x'_0)$}
\pcline{\ind $W_{\mathrm{out}} \leftarrow \log q(X_{\mathrm{fwd}}\given x_0)
  - \log p_\theta(\overleftarrow{X_{\mathrm{fwd}}}\given x_{t^*})$}
\pcline{\ind $W_{\mathrm{in}} \leftarrow \log p_\theta(X_{\mathrm{rev}}\given x_{t^*})
  - \log q(\overleftarrow{X_{\mathrm{rev}}}\given x'_0)$}
\pcline{\ind $W \leftarrow W_{\mathrm{energy}} + W_{\mathrm{out}} + W_{\mathrm{in}}$}
\pcline{\ind \pccomment{4. Metropolis--Hastings step}}
\pcline{\ind Draw $u\sim\mathrm{Uniform}(0,1)$;\quad
  \pckw{if} $u < \min\{1,\exp(-W)\}$ \pckw{then} $x_0\leftarrow x'_0$
  \pckw{end if}}
\pcline{\ind \pckw{if} $\kappa\mid k$ \pckw{then}
  store $x_0^{(k/\kappa)}\leftarrow x_0$ \pckw{end if}}
\pcline{\pckw{end for};\quad \pckw{return} $\{x_0^{(k)}\}$}
\end{pseudocode}

\section{Müller--Brown: setup and additional results}
\label{app:mb-tstar}

All experiments are run at target inverse temperatures $\beta \in \{1.00,\, 1.50,\, 2.00,\, 3.00\}$, with
the main text reporting $t^\ast = 0.02$. Densities and reference occupancies are evaluated on a
$299 \times 299$ quadrature grid over $x \in [-1.5, 1.2]$, $y \in [-0.5, 2.0]$, which contains all
three minima. A basin is defined as the connected sub-level region
$\{x : U(x) \le U_{\min} + \Delta u\}$ containing the corresponding minimum, with
$\Delta u = 0.85$. The reference occupancy $p_{\mathrm{ref}}$ is the
Boltzmann weight integrated over that region by grid quadrature. For the central basin this gives
$p_{\mathrm{ref}} = 3.97 \times 10^{-2}$, $1.22 \times 10^{-2}$, $2.99 \times 10^{-3}$ and $1.39 \times 10^{-4}$ at
$\beta = 1.00$, $1.50$, $2.00$ and $3.00$, respectively, so the mode of interest carries under $4\%$ of the mass even at
the highest temperature.

The two pipelines are matched on total Langevin simulated time, burn-in included. The total budget for restrained data is  $25 + 10{,}000 \times 0.25 = 2{,}525$ time units for  probe generation and
$10{,}000 \times 0.3 = 3{,}000$ on restrained runs, for $5{,}525$ in total. The budget for the baseline model is the
same $5{,}525$ time units, spent on one unrestrained trajectory at the target $\beta$, seeded from a
randomly chosen probe point: $25$ units of burn-in followed by recording at the same $0.25$
interval, which yields $22{,}000$ points. The baseline therefore receives $2.2\times$ more training
data at identical simulation cost.

For the forward diffusion process in equation \eqref{eq:forward}, we use the linear variance-preserving (VP) schedule~\citep{Song2021}:
\begin{equation}
  \alpha_t^2=\exp(-0.1t-9.95t^2),
  \qquad
  \sigma_t^2=1-\alpha_t^2.
  \label{eq:experimental_vp_schedule}
\end{equation}
Both models use Adam~\citep{adam} with learning rate $10^{-3}$, batch size $1{,}000$ and an Exponential Moving Average (EMA)~\citep{ema} with decay $0.99$. Models are trained for $10{,}000$ iterations and
evaluated with EMA weights. At sampling time all Gibbs
chains start from the same point $(-0.558, 1.442)$, so the chains must locate the remaining
basins unaided. Each replicate of the pipeline is run at its own seed, $20$ in total, and the $t^\ast$ sweep
below uses the first five of them.

\begin{figure}[htb]
\centering
\includegraphics[width=0.7\textwidth]{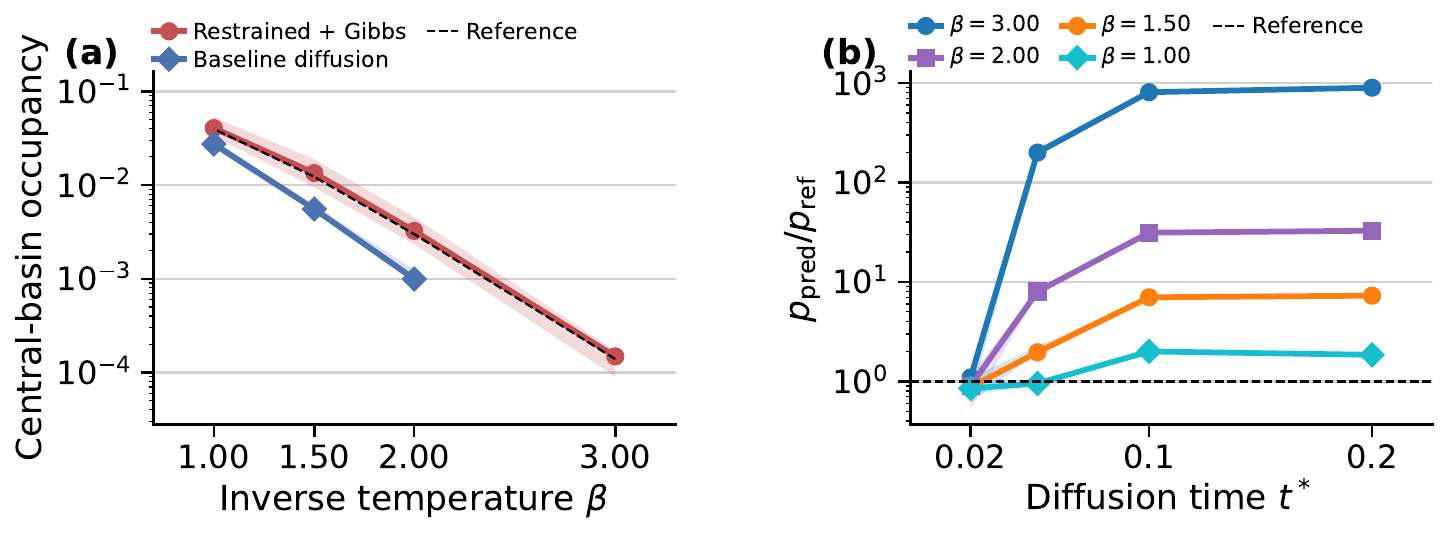}
\caption{
    (a) The raw values of central-basin occupancy plotted in Figure~\ref{fig:mb_results}(c). For $\beta=3$,
    the baseline model gives zero occupancy.
    (b) Central-basin occupancy relative to its exact value,
    $p_{\mathrm{pred}} / p_{\mathrm{ref}}$, against $t^\ast$ at each inverse temperature $\beta$,
    after $10{,}000$ training iterations. Median and interquartile range are calculated over five independently seeded runs.
}
\label{fig:mb-tstar}
\end{figure}

The hyperparameter $t^\ast$ sets the restraint width through
$s = \sigma_{t^\ast} / \alpha_{t^\ast}$, so increasing it weakens the harmonic restraint. Its
influence is most pronounced at large $\beta$, where the target has sharper gradients and taller
barriers. Figure~\ref{fig:mb-tstar}(b) reports the central-basin occupancy relative to its exact
value, $p_{\mathrm{pred}} / p_{\mathrm{ref}}$, against
$t^\ast \in \{0.02,\, 0.05,\, 0.1,\, 0.2\}$ at each $\beta$, with every other parameter held at the
values given in Section~\ref{sec:mueller_brown_exp}; in particular the restrained trajectory length is fixed at
$0.3$ time units for every $t^\ast$, so the simulation budget does not vary along the horizontal
axis. This sweep uses the first five of the seeded replicates.

At fixed restrained-trajectory length the occupancy is close to exact at $t^\ast = 0.02$, with median ratios of $0.85$, $0.87$, $0.90$ and $1.10$ at $\beta = 1.00$, $1.50$,
$2.00$ and $3.00$, respectively. As $t^\ast$ grows the basin is increasingly overpopulated, especially at larger $\beta$.
The restrained density is smoothed into a single well by the harmonic term only when that term dominates. As $t^\ast$
grows the restraint weakens, the barrier structure of $\beta U$ survives in the restrained density,
and sampling training data from it re-introduces the barrier-crossing problem. The fixed-length
trajectories used to generate training data then no longer reach the intended conditional, so the
training pairs are biased and the learned model inherits that bias. Recovering the correct
conditional in this regime would require longer restrained trajectories, or more bridge points per
probe, than at small $t^\ast$.

Small $t^\ast$ smooths the restrained density more strongly and so makes the inner sampling problem
easier, with an important drawback. Successive Gibbs
iterations become strongly correlated, since each sweep perturbs the state only by an amount set by
$t^\ast$ and the effective number of independent samples per iteration therefore falls. In particular, transitions between basins become slower
during sampling. The method thus exhibits a
trade-off between the cost of generating the training data and the efficiency of the resulting sampler.

\section{Alanine dipeptide: setup and additional results}
\label{app:NANMA}

\subsection{System, preprocessing, and reference distribution}

\paragraph{Explicit-solvent molecular dynamics.}
The solvated system contains a 22-atom alanine-dipeptide solute and 628 TIP3P
water molecules in a cubic box of side length $2.7222$\,nm (1906 atoms total).
We use the OpenMM~\citep{openmm} implementations of the AMBER99SB-ILDN protein force field~\citep{amber99} and
TIP3P water, particle-mesh Ewald~\citep{Ewald} electrostatics with a $0.9$\,nm real-space
cutoff, constrained solute hydrogen bonds, rigid water, and center-of-mass
motion removal.  All trajectories use a Langevin-middle integrator~\citep{lfmiddle} with
friction $1\,\mathrm{ps}^{-1}$ and step size $2$\,fs.  After at most 2000
energy-minimization iterations, velocities are drawn at the simulation
temperature and the system is propagated for 10\,ps before the first stored
frame.  Production calculations use single-precision CUDA.

Five independent 300\,K trajectories are generated with seeds 9000--9004 and
saved every 1\,ps.  The reported experiment uses each trajectory over
$0$--$5.05\,\mu\mathrm{s}$, treats the first $50$\,ns as warm-up, and hence has
$25.0\,\mu\mathrm{s}$ of production data and a total reported MD cost of
$25.25\,\mu\mathrm{s}$.  A separate 250\,ns 500\,K trajectory (seed 9500) stores the full solvated state every 10\,ps, supplying the 5000 probe candidates used in Setting~II.  Water is retained during MD but discarded before training.

Each equilibrium solute frame is Kabsch-aligned, using all 22 solute atoms, to a fixed reference structure used consistently across all experiments. The
restrained-MD preprocessing described below additionally makes each solute
whole under periodic boundary conditions.  Coordinates are expressed in nm
and multiplied by 20 before entering the model.  Baseline model training uses a
50\,ps subsample of the production interval.

\paragraph{Ramachandran reference and basins.}
The reference histogram uses all 25,000,000 raw 1\,ps frames in
$(0.05,5.05]\,\mu\mathrm{s}$ from the five 300\,K trajectories.  We form a
$300\times300$ periodic histogram over $[-\pi,\pi)^2$, smooth its density with
a wrapped Gaussian of standard deviation two bins, and define the
dimensionless free energy $F=-\log p+\mathrm{const}$.  We identify the five
distinct local minima listed in Table~\ref{tab:NANMA_basins}.  Basin $R_i$ is the
four-neighbor, periodically connected component containing minimum $i$ and
satisfying $F-F_i\leq\Delta F_i$.  This construction gives
$p_{\mathrm{ref}}(R_5)=0.001574$ ($0.1574\%$).

\begin{table}[H]
\centering
\small
\begin{tabular}{
  c
  S[table-format=-1.4]
  S[table-format=-1.4]
  S[table-format=1.1]
}
\toprule
{Basin} & {minimum $\phi$ (rad)} & {minimum $\psi$ (rad)} & {$\Delta F/k_{\mathrm B}T$} \\
\midrule
$R_1$ & -2.5237 &  2.7541 & 0.6 \\
$R_2$ & -1.2462 &  2.6704 & 1.0 \\
$R_3$ & -1.3090 & -0.3037 & 1.5 \\
$R_4$ &  0.9320 &  0.5341 & 3.0 \\
$R_5$ &  1.0367 &  3.0264 & 2.0 \\
\bottomrule
\end{tabular}
\caption{Local-minimum coordinates and basin-specific free-energy barriers used
for all alanine dipeptide analyses.}
\label{tab:NANMA_basins}
\end{table}

\subsection{Restrained-pair construction}

\paragraph{Shared Cartesian CV.}
Let $A\in\mathbb{R}^{30\times66}$ select the Cartesian coordinates of the ten
solute heavy atoms with zero-based atom indices
$(1,4,5,6,8,10,14,15,16,18)$.  We use the same VP schedule \eqref{eq:experimental_vp_schedule}  for  the diffusion model. At $t^*=0.40$, $(\alpha_{t^*},\sigma_{t^*})=(0.4421969,0.8969180)$ and the
matched restraint width is
$s=\sigma_{t^*}/\alpha_{t^*}=2.028323$ in scaled model coordinates, or
$0.101416$\,nm in physical coordinates.

\paragraph{Setting I: synthetic pairs.}
For each MD-budget level, only its matching equilibrium pool is used.  The
Ramachandran plane is divided into a $50\times50$ grid.  If an occupied cell
contains $n_{ij}$ frames, its probe quota is proportional to
$n_{ij}^{0.5}$, with integer remainders assigned by largest fractional part
and quotas capped by cell occupancy.  We then draw 10,000 distinct equilibrium
probes without replacement within their cells.  For each training run and MD-budget level, the 10,000 probes and their posterior-matched synthetic pairs are independently resampled using seed-specific random streams.

Each clean probe produces 30 pairs. In scaled coordinates, a target CV is
drawn as
\[
y=A\tilde{x}+s\xi,
\qquad
\xi\sim\mathcal{N}(0,AA^\top).
\]
A source frame \(x_0\) is drawn by discrete resampling from the linear-CV
marginal in Equation~\eqref{eq:cv_x0_marginal}, with the endpoint target in
that equation set to \(\alpha_{t^*}y\). Equivalently, its unnormalized
resampling weight is
\[
w(x_0\mid y)
\propto
\exp\!\left[
-\frac{1}{2s^2}
(Ax_0-y)^\top(AA^\top)^{-1}(Ax_0-y)
\right].
\]
Given \(x_0\), the conditional distribution in
Equation~\eqref{eq:cv_xt_conditional}, again evaluated at endpoint target
\(\alpha_{t^*}y\), gives the normalized endpoint
\[
z=\frac{x_{t^*}}{\alpha_{t^*}}
=
A^\top(AA^\top)^{-1}y
+
\left[I-A^\top(AA^\top)^{-1}A\right](x_0+s\epsilon),
\qquad
\epsilon\sim\mathcal{N}(0,I).
\]
Therefore \(Az=y\) exactly. Every budget contains 300,000 synthetic pairs and
requires no additional MD or energy evaluations. The equilibrium-pool mean is
subtracted before lower-time training and restored after sampling.

\paragraph{Setting II: restrained-MD pairs.}
The 250\,ns of the 500\,K trajectory is sampled every 50\,ps, giving
5000 candidates that include the complete solvent state and periodic box.
For each candidate, Gaussian noise of standard deviation $0.101416$\,nm is
added only to the ten selected heavy atoms to define a restraint center.
Starting from the corresponding solvated state, velocities are redrawn at
300\,K and a 5\,ns trajectory is run without minimization using the same
physical model as above. We implement the harmonic factor in
Equation~\eqref{eq:cv_x0_marginal} in physical coordinates using
minimum-image displacements:
\[
U_{\mathrm{rest}}(x)
=
\frac{k}{2}\sum_{a\in\mathcal H}
\operatorname{periodicdistance}(x_a,z_a)^2,
\qquad
k=\frac{RT}{s^2}
=242.52\;\mathrm{kJ\,mol^{-1}\,nm^{-2}}.
\]
Frames are stored every 10\,ps, but only the 30 frames at
2.1,2.2,\ldots,5.0\,ns are retained.  Before pair construction, each solute is
made whole, the restraint displacement is taken by the minimum-image
convention, and the configuration and target share the same framewise Kabsch
transform.  The unconstrained component of $z$ is then resampled analytically
as above, which preserves $Az=y$ exactly. The full dataset contains 150,000 pairs, and its centering shift is the mean of the 5000 aligned high-temperature candidates.

\subsection{Matched MD budgets and repeated runs}

For the MD-budget curve, all five equilibrium trajectories are retained and
their production intervals are shortened together.  The restrained method
uses a matching number of 5\,ns trajectories, so both sides have cost
$5L+0.25\,\mu\mathrm{s}$ at level $L$ (Table~\ref{tab:NANMA_budgets}).  The
25.0\,$\mu$s reference distribution is used to evaluate every level and is
not treated as training data at smaller budgets.

\begin{table}[H]
\centering
\small
\setlength{\tabcolsep}{4pt}
\begin{tabular}{@{}ccccc@{}}
\toprule
&
\multicolumn{2}{c}{Baseline model training dataset}
&
\multicolumn{2}{c}{RDB model training dataset}
\\
\cmidrule(lr){2-3}
\cmidrule(lr){4-5}
\shortstack{Matched MD\\budget ($\mu$s)}
&
\shortstack{Production interval\\per trajectory ($\mu$s)}
&
\shortstack{Training\\configurations}
&
\shortstack{Restrained-MD\\probes}
&
\shortstack{Retained training\\pairs}
\\
\midrule
 5.25 & $(0.05,1.05]$ & 100,000 & 1,000 &  30,000 \\
10.25 & $(0.05,2.05]$ & 200,000 & 2,000 &  60,000 \\
15.25 & $(0.05,3.05]$ & 300,000 & 3,000 &  90,000 \\
20.25 & $(0.05,4.05]$ & 400,000 & 4,000 & 120,000 \\
25.25 & $(0.05,5.05]$ & 500,000 & 5,000 & 150,000 \\
\bottomrule
\end{tabular}
\caption{
Comparison of the two model-training datasets at matched MD budgets.
The baseline model dataset pools configurations from five equilibrium trajectories
after discarding the first 50\,ns of each trajectory and subsampling the
remaining production segments every 50\,ps. The restrained-MD RDB model dataset is
constructed from a 250\,ns high-temperature probe trajectory and matched
restrained simulations. After discarding the equilibration portion of each
restrained trajectory, 30 target configurations are retained per probe.
Consequently, the two approaches use the same total MD time but different
numbers of retained training configurations.
}
\label{tab:NANMA_budgets}
\end{table}

We use 20 runs with training seeds 42--61.  Base training uses the same equilibrium pool across seeds. For synthetic-pair refinement, the probes and posterior-matched pairs are independently resampled for every training seed at each MD-budget level. For restrained MD we uniformly select probes without replacement. At the maximum budget all runs use all 5000 probes.

\subsection{Model, optimization, and sampling}

The denoiser treats the 22 atoms as tokens.  A two-layer GELU~\citep{gelus} MLP maps each
3-vector to width 128; learned atom-slot and atomic-number embeddings and a
sinusoidal time embedding followed by a two-layer GELU MLP are added.  Three
pre-norm Transformer~\citep{attention} encoder layers use eight attention heads, feed-forward
width 512, GELU activations, and zero dropout.  A linear head predicts one
three-dimensional noise vector per atom.

All models use batch size 4096, Adam with constant learning rate $10^{-3}$,
and an EMA with decay 0.99. The baseline model is trained from scratch for
50,000 updates with diffusion time $t$ uniform on
$[10^{-4},1-10^{-4}]$.  The synthetic-pair RDB refinement initializes both the live and
EMA models from the base-20{,}000 EMA checkpoint, resets the optimizer and update
counter, and performs 30,000 bridge updates with diffusion time $t$ uniform on
$[10^{-4},t^*]$; no equilibrium examples are mixed into these updates.  The
restrained-MD RDB model is initialized randomly and trained from scratch for
50,000 lower-time bridge updates.  The training-budget curve saves and
evaluates checkpoints every 2000 updates.

Each evaluation uses 100,000 generated solute configurations in two batches of
50,000.  Euler--Maruyama reverse sampling uses  $\frac{1}{300}$ time step for baseline model and $\frac{t^*}{100}$ time step for synthetic-pair RDB model. Baseline sampling runs the 50{,}000 model from standard Gaussian noise over $1\to0$.  Sampling of the synthetic-pair RDB model
 integrates the baseline 20{,}000 model over $1\to t^*$, shifts to the
centered coordinate system, and runs the RDB model over $t^*\to0$.
Restrained-MD RDB model evaluation initializes 100,000 Gaussian states at $t^*$ and
runs one $t^*\to0$ reverse pass; this is Gibbs iteration 0.  Each of the
next 50 iterations applies the analytic forward update
$x_{t^*}=\alpha_{t^*}x_0+\sigma_{t^*}\epsilon$ followed by a 100-point reverse
pass.  Reported restrained-MD RDB metrics are the arithmetic average of the ten
per-iteration metrics at iterations 41--50. PyTorch deterministic algorithms are enabled, cuDNN benchmarking is disabled.

\subsection{Metrics and additional results}

For every generated set, $(\phi,\psi)$ is accumulated on the same
$300\times300$ grid as the reference.  With $m=(p+q)/2$, the reported global
metric is
\begin{equation}
 d_{\mathrm{JS}}(p,q)=
 \left[\tfrac12\mathrm{KL}_2(p\Vert m)+
       \tfrac12\mathrm{KL}_2(q\Vert m)\right]^{1/2},
 \label{eq:js_distance}
\end{equation}
where zero-count terms are omitted.  The rare-state relative error is
$p_{\mathrm{gen}}(R_5)/p_{\mathrm{ref}}(R_5)-1$.  Figure~\ref{fig:NANMA_results}
reports medians across the 20 runs; bands and
error bars span the first and third quartiles.  For the restrained MD RDB model, the
last-ten-iteration average described above is computed within each run before
the across-run median and quartiles are taken.

\begin{figure}[t]
\centering
\includegraphics[width=\textwidth]{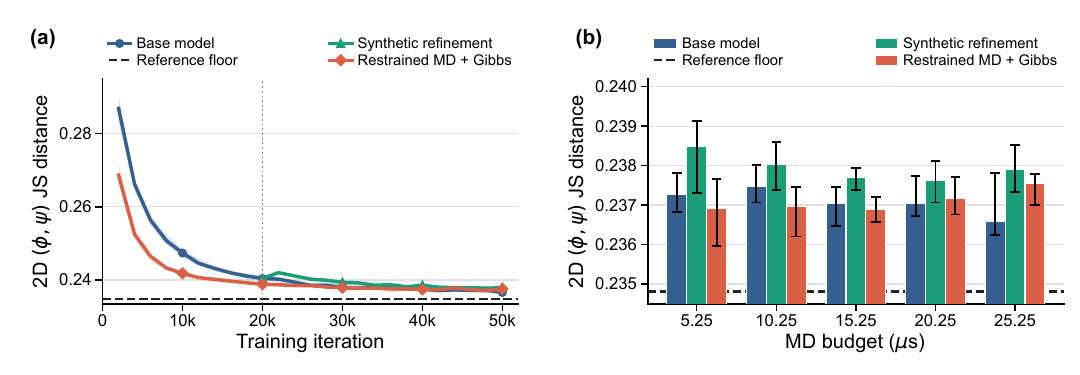}
\caption{
Two-dimensional Ramachandran JS distance versus (a) training updates
at a fixed $25.25\,\mu$s MD budget and (b) matched MD budget at 50{,}000 updates.
Curves and bars show medians over 20 training runs, and bands and error bars show
interquartile ranges.  Dashed lines indicate the expected JS distance between
a 100,000-sample multinomial draw and the 25.0\,$\mu$s reference histogram.
}
\label{fig:NANMA_js_trends}
\end{figure}

Because the generated histogram contains only 100,000 samples spread over
90,000 bins, even an exact generator has nonzero empirical JS distance.  We
estimate this finite-sample floor by 5000 independent multinomial draws of
100,000 samples from the 25.0\,$\mu$s reference histogram; its mean is
0.234812.  Figure~\ref{fig:NANMA_js_trends} shows that all three methods approach
this floor during training and remain close to it throughout the MD-budget
sweep, supporting the conclusion that the improvement in $R_5$ does not come
at the expense of global fidelity.

\section{Chignolin: setup and additional results}
\label{app:chig}

\subsection{Reference simulation and collective variables}

We simulate CLN025 (Tyr-Tyr-Asp-Pro-Glu-Thr-Gly-Thr-Trp-Tyr) at 340\,K with OpenMM using the AMBER14 ff14SB protein force field~\citep{amberff14sb} and GBN2 implicit-solvent model~\citep{gbn2} using the protein.ff14SB.xml and gbn2.xml definitions supplied with OpenMM.  The system
uses no nonbonded cutoff, constrains
bonds to hydrogen, removes center-of-mass motion, and sets the solute and
solvent dielectric constants to 1 and 78.5.  A Langevin-middle integrator uses
a 2\,fs step and friction $1\,\mathrm{ps}^{-1}$.  Velocities are initialized
at 340\,K with seed 2026061800.  The trajectory is propagated for
$22.36\,\mu$s and stored every 10\,ps, yielding 2,236,000 frames.

Only the 93 heavy atoms are retained.  Coordinates are converted to
\AA{} and each frame is Kabsch-aligned, using the ten C$\alpha$ atoms, to a fixed folded reference structure used consistently across all experiments.  The two evaluation
coordinates are
\begin{equation}
 \mathrm{NO1}=d(\mathrm{Asp3\,N},\mathrm{Gly7\,O}),\qquad
 \mathrm{NO2}=d(\mathrm{Asp3\,N},\mathrm{Thr8\,O}).
\end{equation}
The folded region shown in Figure~\ref{fig:chignolin}(a) is
$4.5<\mathrm{NO1}<8.0$\,\AA{} and $\mathrm{NO2}<3.6$\,\AA{}.  The
misfolded region used for every quantitative result is
\begin{equation}
 \mathcal R_{\mathrm{mis}}=
 \{\mathrm{NO1}<3.7\,\text{\AA},\ 5.0<\mathrm{NO2}<8.0\,\text{\AA}\}.
\end{equation}

\subsection{Data budgets and fixed reference}

The data-scaling experiment uses only the first $22\,\mu$s.  We discard
the first $2\,\mu$s as warm-up and define a single reference distribution
from the remaining 2,000,000 frames in $(2,22]\,\mu$s.  This reference
contains 7494 misfolded frames, hence
$p_{\mathrm{ref}}(\mathcal R_{\mathrm{mis}})=0.003747$.  All generated
samples, including models trained at smaller data budgets, are compared with
this same reference.

The four training windows are prefixes of the post-warm-up interval
(Table~\ref{tab:chig_data_windows}).  For each window we recompute the
per-atom Cartesian center and one global scale, the square root of the mean
coordinate variance.  No frame after the stated endpoint contributes to its
normalization, autoencoder, latent statistics, probes, or diffusion training.

\begin{table}[H]
\centering
\small
\begin{tabular}{@{}ccc@{}}
\toprule
MD budget ($\mu$s) & Source interval ($\mu$s) & Frames \\
\midrule
 7  & $(2,7]$  &   500,000 \\
12  & $(2,12]$ & 1,000,000  \\
17  & $(2,17]$ & 1,500,000 \\
22  & $(2,22]$ & 2,000,000  \\
\bottomrule
\end{tabular}
\caption{Chignolin training windows.  The 2--22\,$\mu$s reference is fixed
across all four data budgets.}
\label{tab:chig_data_windows}
\end{table}

\subsection{Autoencoder and latent representation}

We train one independent autoencoder for each data budget and share that
representation across all ten diffusion seeds at that budget. The encoder and decoder each have width 512 and four residual blocks.  A block applies layer normalization followed by two linear layers with a SiLU
activation and adds the result to the skip connection with a
$1/\sqrt{2}$ factor.  The bottleneck has dimension 128.

With coordinates in their standardized units and distances in \AA{}, let
\(\tilde{\mathbf r}_{na}\) and \(\hat{\tilde{\mathbf r}}_{na}\) denote the
input and reconstructed Cartesian coordinates of atom \(a\) in configuration
\(n\), and let \(\mathbf r_{na}\) and \(\hat{\mathbf r}_{na}\) be the
corresponding coordinates after transforming back to \(\text{\AA}\). For a
batch of \(B\) configurations, the four loss terms are
\[
\mathcal L_{\mathrm{coord}}
=
\frac{1}{279B}
\sum_{n=1}^{B}\sum_{a=1}^{93}
\left\|
\hat{\tilde{\mathbf r}}_{na}
-
\tilde{\mathbf r}_{na}
\right\|_2^2 ,
\]
\[
\mathcal L_{\mathrm{bond}}
=
\frac{1}{98B}
\sum_{n=1}^{B}
\sum_{(a,b)\in\mathcal E_{\mathrm{bond}}}
\left(
\|\hat{\mathbf r}_{na}-\hat{\mathbf r}_{nb}\|_2
-
\|\mathbf r_{na}-\mathbf r_{nb}\|_2
\right)^2 ,
\]
\[
\mathcal L_{\mathrm{C\alpha}}
=
\frac{1}{45B}
\sum_{n=1}^{B}
\sum_{(a,b)\in\mathcal E_{\mathrm{C\alpha}}}
\left(
\|\hat{\mathbf r}_{na}-\hat{\mathbf r}_{nb}\|_2
-
\|\mathbf r_{na}-\mathbf r_{nb}\|_2
\right)^2 ,
\]
and
\[
\mathcal L_{\mathrm{NO}}
=
\frac{1}{2B}
\sum_{n=1}^{B}\sum_{k=1}^{2}
\left(\hat d_{nk}^{\mathrm{NO}}-d_{nk}^{\mathrm{NO}}\right)^2 .
\]
Here, \(\mathcal E_{\mathrm{bond}}\) contains the 98 covalent heavy-atom bonds in the fixed molecular topology, while \(\mathcal E_{\mathrm{C\alpha}}\) contains all \(\binom{10}{2}=45\) pairwise distances among the ten C\(\alpha\) atoms. The final term explicitly preserves NO1 and NO2, the two distances used to distinguish the folded and misfolded conformations. The complete objective is
\[
\mathcal L_{\mathrm{AE}}
=
\mathcal L_{\mathrm{coord}}
+0.1\mathcal L_{\mathrm{bond}}
+0.1\mathcal L_{\mathrm{C\alpha}}
+\mathcal L_{\mathrm{NO}} .
\]
Thus, the coordinate term controls overall reconstruction, the bond term preserves local molecular geometry, the C\(\alpha\) term preserves the global backbone arrangement, and the NO term emphasizes reconstruction accuracy in the rare-state collective variables.

We use Adam, batch size 1024, gradient clipping
at 1, and bfloat16 autocasting.  The learning rate warms linearly to
$2\times10^{-4}$ over 2000 updates and follows a cosine schedule with
100,000-update horizon and minimum multiplier 0.1.  Training stops at the
fixed 60,000-update endpoint; the common autoencoder seed is 2026090744.
Across the four validation splits, the coordinate RMSD is
$0.0406$--$0.0420$\,\AA{} and the decoded-to-original misfolded-population
ratio is $0.983$--$1.008$.

After encoding every frame, each latent coordinate is standardized.  Its mean
and standard deviation are estimated from at most 200,000 evenly spaced
frames in the corresponding training window.  All posterior construction,
diffusion training, and reverse sampling below occur in this standardized
128-dimensional latent space.

\subsection{Probe allocation and synthetic posterior pairs}

For each MD budget, the $(\mathrm{NO1},\mathrm{NO2})$ plane of that training
window is divided into a $50\times50$ grid.  If cell $b$ contains $n_b$
frames, 10,000 distinct probe frames are allocated in proportion to
$n_b^{0.5}$; integer remainders are assigned by largest fractional part.
Sampling within cells is uniform without replacement.

We use the same linear VP schedule as
Equation~\eqref{eq:experimental_vp_schedule}.
Let $\tilde z_j$ be a standardized latent probe and
$s_{t^*}=\sigma_{t^*}/\alpha_{t^*}$.  We draw a perturbed restraint center
$y_j=\tilde z_j+s_{t^*}\xi_j$, where
$\xi_j\sim\mathcal N(0,I)$. For each $y_j$, we sample 30 clean endpoints
with replacement from the empirical version of the restrained distribution
in Equation~\eqref{eq:restrained}, taking $z=y_j$ and $s=s_{t^*}$ and
assigning equal base mass to every latent frame.
This produces 300,000 bridge pairs per data
budget without additional MD.
We evaluate $t^*=0.30$ and $0.35$, for which
$(\alpha_{t^*},\sigma_{t^*},s_{t^*})$ equals
$(0.62955,0.77696,1.23415)$ and
$(0.53423,0.84534,1.58237)$.  A pair set is shared across the ten diffusion
seeds at fixed data budget and $t^*$.

\subsection{Model training and inference}

The denoiser is a time-conditioned residual MLP with width 1024, eight FiLM~\citep{film}
residual blocks, and a 256-dimensional Fourier time embedding~\citep{attention}.  It predicts
VP noise in the standardized latent space.  We use seeds 42--51, batch size
2048, Adam, EMA decay 0.99, gradient clipping at 1, and bfloat16 autocasting.
The learning-rate schedule warms to $2\times10^{-4}$ over 2000 updates and
then decays by the same cosine rule to 0.1 of its peak at 160,000 updates.

The baseline model is trained for 160,000 updates with diffusion time
$t\sim\mathrm{Uniform}(10^{-4},1-10^{-4})$.  Synthetic-pair RDB refinement copies the
complete baseline model, EMA, optimizer, and random state at the 80,000-update
checkpoint and performs 80,000 additional bridge updates with diffusion time
$t\sim\mathrm{Uniform}(10^{-4},t^*)$.  The optimizer and global learning-rate
clock are not reset, and no equilibrium examples are mixed into these lower-time
updates.  The $t^*=0.30$ runs reuse the audited autoencoders and base
trajectories from the $t^*=0.35$ experiment; exact baseline 80{,}000 and baseline 160{,}000 checkpoint
state hashes are checked before reuse.

Every evaluation generates 100,000 configurations with the EMA parameters.
Euler--Maruyama sampling uses one common 300-transition grid.  Baseline model sampling
uses baseline 160{,}000 from $1$ to $0$.  Synthetic-pair RDB model sampling uses baseline 80{,}000
from $1$ to $t^*$ and the RDB refined model from $t^*$ to $0$.  The same
evaluation random stream is used for baseline and both synthetic-pair RDB model branches within
each seed, yielding paired comparisons.  Training curves evaluate every
10,000 updates; the smaller data budgets retain only the final 160,000-update
evaluation.

\subsection{Metrics and bridge-time sensitivity}
\label{sec:metrics}

The fixed 2--22\,$\mu$s reference defines 100 interior bins along each NO
coordinate, with endpoints at its $10^{-4}$ and $0.9999$ quantiles.  Explicit
underflow and overflow bins give a $102\times102$ probability table.  We
compare the generated and reference tables using the base-2 JS distance in
Equation~\eqref{eq:js_distance}. The expected finite-sample JS distance is $0.09822$, estimated from 200 independent 100,000-sample resamples
of the fixed $2$--$22\,\mu\mathrm{s}$ reference ensemble.
Rare-state accuracy is the generated fraction
$p_{\mathrm{gen}}(\mathcal R_{\mathrm{mis}})$.  Figures and tables report the
median over ten independently trained diffusion models; bands and error bars
span the first and third quartiles.

\begin{figure}[htb]
\centering
\includegraphics[width=\textwidth]{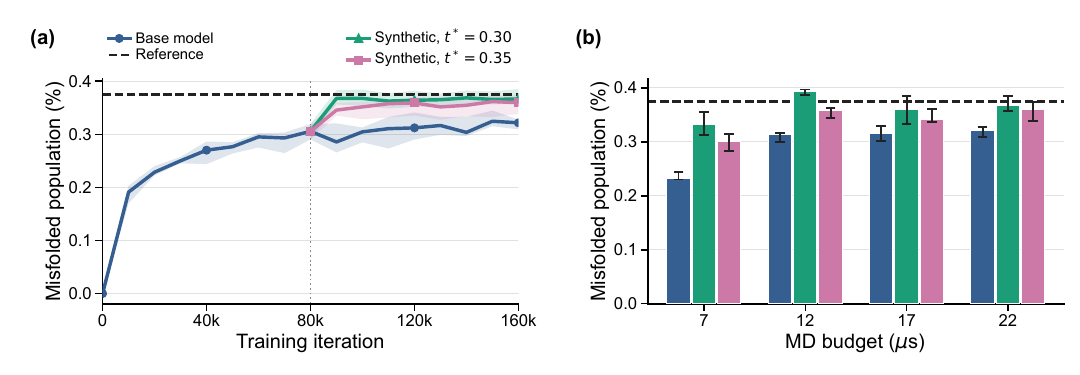}
\caption{
Chignolin bridge-time comparison.  Generated misfolded population
versus (a) training iteration at a fixed 22\,$\mu$s MD budget and
(b) MD budget at 160{,}000 total updates.  Curves and bars show
medians over ten seeds; bands and error bars show interquartile ranges.  The
dashed line is the fixed reference population, $0.3747\%$, and the dotted
line in (a) marks the 80{,}000 branch point.
}
\label{fig:chig_tstar_sensitivity}
\end{figure}

\begin{figure}[htb]
\centering
\includegraphics[width=\textwidth]{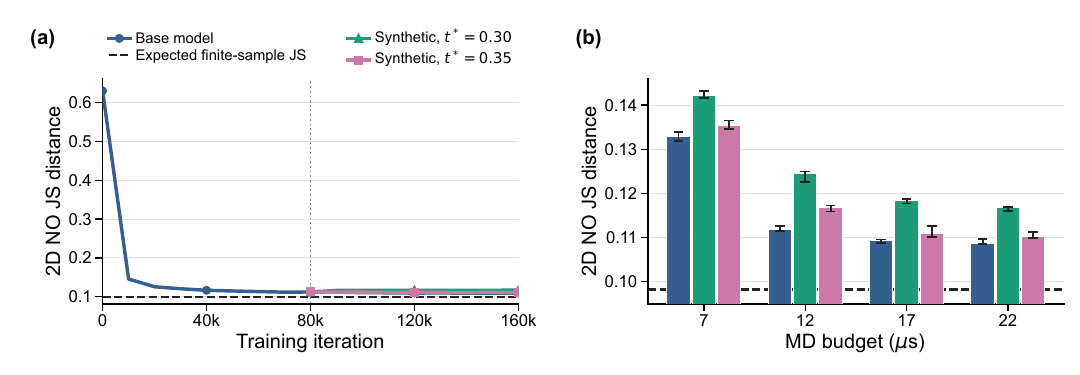}
\caption{
Chignolin global fidelity. Two-dimensional NO JS distance versus
(a) training iteration at a fixed $22\,\mu$s MD budget and
(b) MD budget at 160{,}000 total updates, for both
$t^*=0.30$ and $0.35$. Curves and bars show medians over ten seeds;
bands and error bars show interquartile ranges. Dashed lines indicate the expected finite-sample JS distance,
$0.09822$, estimated from 200 independent 100,000-sample resamples
of the fixed $2$--$22\,\mu\mathrm{s}$ reference ensemble. The dotted
line in (a) marks the 80{,}000 branch point.
}
\label{fig:chig_js_results}
\end{figure}

\begin{table}[H]
\centering
\small
\setlength{\tabcolsep}{2.5pt}
\resizebox{\textwidth}{!}{%
\begin{tabular}{@{}cclccc@{}}
\toprule
MD budget ($\mu$s) & Method &
$p_{\mathrm{gen}}(\mathcal R_{\mathrm{mis}})$ (\%) &
$|\mathrm{relative\ error}|$ (\%) &
\multicolumn{2}{c}{2D NO JS distance} \\
\cmidrule(lr){3-3}\cmidrule(lr){5-6}
& & Median [Q1, Q3] & & Median & [Q1, Q3] \\
\midrule
7  & Base                 & 0.2330 [0.2302, 0.2440] & 37.8 & 0.13290 & [0.13179, 0.13386] \\
   & Synthetic, $t^*=.30$ & 0.3320 [0.3132, 0.3558] & 11.4 & 0.14217 & [0.14167, 0.14316] \\
   & Synthetic, $t^*=.35$ & 0.3010 [0.2833, 0.3145] & 19.7 & 0.13552 & [0.13453, 0.13655] \\
\addlinespace
12 & Base                 & 0.3130 [0.2993, 0.3158] & 16.5 & 0.11184 & [0.11153, 0.11268] \\
   & Synthetic, $t^*=.30$ & 0.3930 [0.3862, 0.3970] &  4.9 & 0.12443 & [0.12254, 0.12498] \\
   & Synthetic, $t^*=.35$ & 0.3580 [0.3448, 0.3628] &  4.5 & 0.11679 & [0.11591, 0.11726] \\
\addlinespace
17 & Base                 & 0.3160 [0.3020, 0.3287] & 15.7 & 0.10946 & [0.10880, 0.10956] \\
   & Synthetic, $t^*=.30$ & 0.3600 [0.3330, 0.3852] &  3.9 & 0.11842 & [0.11772, 0.11872] \\
   & Synthetic, $t^*=.35$ & 0.3420 [0.3365, 0.3615] &  8.7 & 0.11103 & [0.11016, 0.11268] \\
\addlinespace
22 & Base                 & 0.3215 [0.3090, 0.3277] & 14.2 & 0.10875 & [0.10850, 0.10968] \\
   & Synthetic, $t^*=.30$ & 0.3665 [0.3572, 0.3852] &  2.2 & 0.11673 & [0.11599, 0.11690] \\
   & Synthetic, $t^*=.35$ & 0.3595 [0.3382, 0.3732] &  4.1 & 0.11016 & [0.10981, 0.11129] \\
\bottomrule
\end{tabular}}
\caption{Final Chignolin metrics at 160{,}000 total updates.  Relative error is
computed from each across-seed median and the fixed reference population
$0.3747\%$.}
\label{tab:chig_final_metrics}
\end{table}

Both bridge times improve the median misfolded population over the baseline model
at every data budget, but their quantitative tradeoffs differ.
$t^*=0.30$ is usually closer in rare-state population, while $t^*=0.35$
usually incurs a smaller increase in 2D NO JS distance.  At 22\,$\mu$s, for
example, the absolute relative population error is $2.2\%$ at $t^*=0.30$ and
$4.1\%$ at $t^*=0.35$, whereas the median JS distances are 0.11673 and
0.11016. Both nearby choices consistently
improve the rare-state estimate, so performance is not sharply sensitive over
the tested range; however, we do not yet have a general strategy for choosing
the optimal $t^*$.

\end{document}